\documentclass[10pt,twocolumn,letterpaper]{article}
\PassOptionsToPackage{table}{xcolor}

\usepackage[pagenumbers]{cvpr} % To force page numbers, e.g. for an arXiv version

\usepackage{cuted}
\usepackage{caption}

\usepackage{booktabs}
\usepackage{tabularx}
\usepackage{amssymb}
\usepackage{xcolor}

\definecolor{cvprblue}{rgb}{0.21,0.49,0.74}
\usepackage[pagebackref,breaklinks,colorlinks,allcolors=cvprblue]{hyperref}

\def\confName{CVPR}
\def\confYear{2027}

\title{PhysPPF: Learning Robotic Manipulation with Past, Present, and Future Physical Representations}

\author{
Junjie Li$^{1,*}$ \quad
Dawei Yan$^{1,*}$ \quad
Guoting Wei$^{1,*}$ \quad
Ruike Song$^{1}$ \quad
Lu Wang$^{2}$\\
Qingguo Chen$^{2}$ \quad
Zhenbo Luo$^{3}$ \quad
Haokui Zhang$^{1,\dagger}$ \quad
Chunhua Shen$^{4,\dagger}$\\[0.4em]
{\small
$^{1}$Northwestern Polytechnical University
}\\
{\small
$^{2}$Alibaba Cloud Computing \qquad
$^{3}$Xiaomi Inc \qquad
$^{4}$Zhejiang University
}\\[0.3em]
{\tt\small
junjiel@mail.nwpu.edu.cn \quad
ydw4396@mail.nwpu.edu.cn \quad
hkzhang@nwpu.edu.cn
}\\
{\tt\small
wscmyjy@gmail.com \quad
qingguo.cqg@alibaba-inc.com
}\\
{\tt\small
luozhenbo@xiaomi.com \quad
chunhuashen@zju.edu.cn
}\\[0.3em]
{\small
$^{*}$These authors contributed equally.
\quad
$^{\dagger}$Corresponding authors.
}
}

\begin{document}

\maketitle

\setlength{\stripsep}{0pt}

\begin{strip}
    \centering
    \includegraphics[width=1\linewidth]{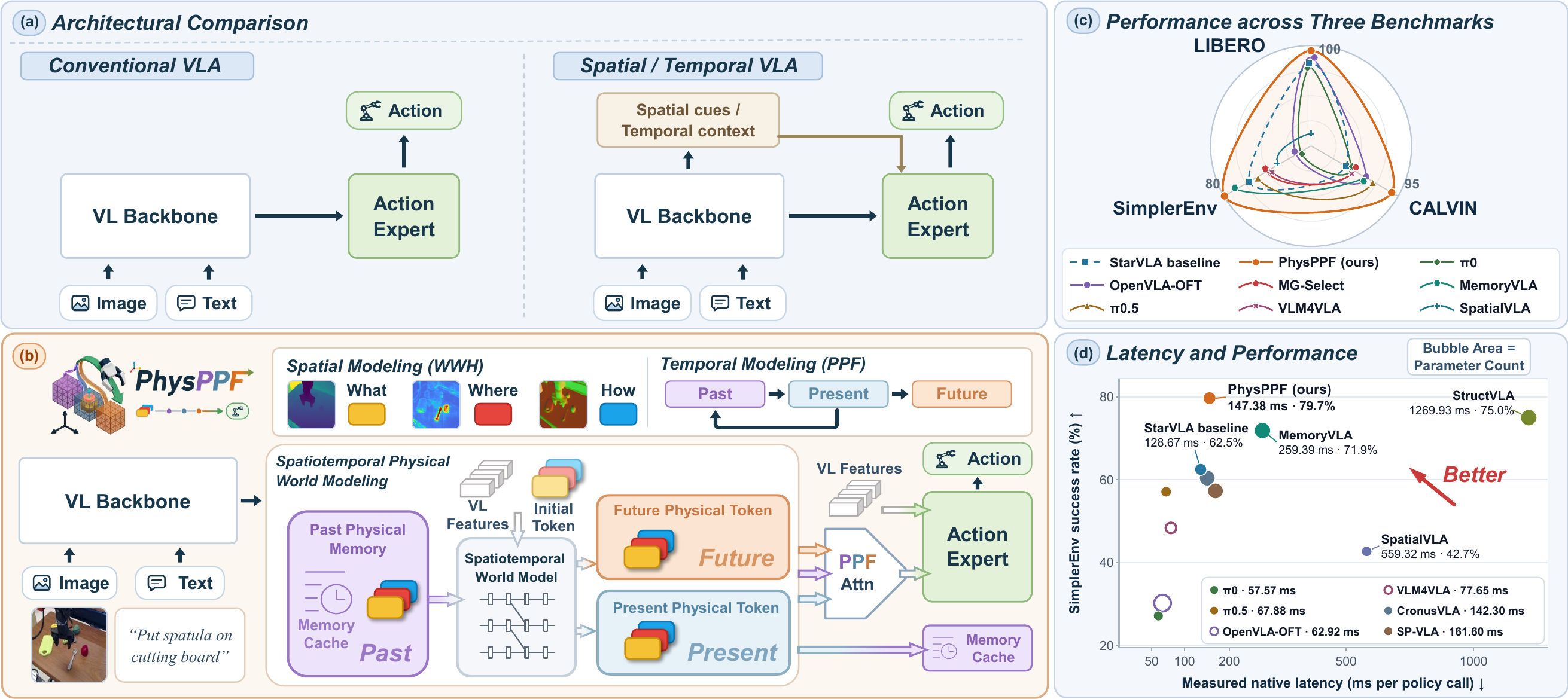}
    % \vspace{-10pt}
    % \vspace{-0.5cm}
    \captionsetup{hypcap=false}
    \captionof{figure}{\textbf{Overview of PhysPPF.} \textbf{(a)} Conventional VLAs rely on single-frame 2D features, while recent works augment spatial cues or temporal context separately. \textbf{(b)} \textbf{PhysPPF} performs unified spatiotemporal modeling in a compact physical token space, integrating \textbf{WWH} (\emph{What-Where-How}) 3D perception and \textbf{PPF} (\emph{Past-Present-Future}) state evolution to guide the action expert. \textbf{(c)} It outperforms prior VLAs across three benchmarks while \textbf{(d)} maintaining modest latency and parameter overhead on SimplerEnv.}
    \label{fig:teaser}
    \captionsetup{hypcap=true}
    % \par\vspace{10pt}
\end{strip}

\begin{abstract}
Vision-language-action (VLA) models have advanced robotic manipulation, yet complex tasks requiring both precise 3D contact and long-horizon coordination remain challenging.
Most existing studies enhance spatial perception or temporal reasoning in VLAs separately.
Spatial methods primarily inject geometric or affordance priors into current-frame perception without tracking multi-step transitions, while history-aware and world-model VLAs typically capture past context or future dynamics in 2D feature or pixel spaces, either lacking explicit 3D contact cues or incurring high generation overhead.
To address this, we propose \textbf{PhysPPF}, a VLA framework that performs unified spatiotemporal modeling in a compact physical token space.
Spatially, \textbf{WWH} (\emph{What-Where-How}) distills pretrained perception teachers exclusively during training into compact token groups capturing \emph{what} 3D structures occupy the scene, \emph{where} to interact, and \emph{how} the gripper should approach, without test-time 3D encoders.
Temporally, operating directly in this WWH space without pixel reconstruction, \textbf{PPF} (\emph{Past-Present-Future}) consolidates executed states into a bounded \emph{past} physical memory and jointly infers \emph{present} and \emph{future} physical tokens via an asymmetric world model.
Parallel gated branches then inject these past, present, and future representations into a flow-matching action expert to guide manipulation without cross-temporal interference.
Extensive experiments on SimplerEnv, CALVIN, and LIBERO show that \textbf{PhysPPF} consistently outperforms existing spatial, history-aware, and world-model VLAs with modest latency and parameter overhead, improving over the StarVLA baseline by +17.2 points on SimplerEnv, +29.3 points on CALVIN, and +3.5 points on LIBERO.
\end{abstract}
\section{Introduction}
\label{sec:introduction}

Enabling robots to perform diverse manipulation skills in physical environments is a central goal in robot learning~\cite{zeng2021transporter,jang2022bcz,shridhar2022peract,brohan2022rt1,jiang2022vima,chi2023diffusionpolicy,zhao2023act,openx2023rtx,octo2024}.
Recent vision-language-action (VLA) models advance this direction by leveraging pretrained vision-language (VL) backbones to generate actions from visual observations and language instructions~\cite{brohan2023rt2,kim2024openvla,black2024pi0,black2025pi05,nvidia2025grootn1,kim2025openvlaoft,zhang2026atomicvla,community2026starvla}.
Despite this progress, reliable execution remains challenging for complex tasks requiring both precise physical contact and long-horizon coordination.
Consider the instruction \emph{``open the drawer and place the object inside''}~\cite{mees2022calvin,li2026cronusvla}: the robot must first align its gripper with the handle to pull the drawer open, then pick up the object while remembering that the drawer is already open, and finally place the object inside without collision.
However, conventional VLAs pass single-frame 2D features from the VL backbone directly to the action expert (Fig.~\ref{fig:teaser}(a), left), lacking both 3D spatial cues for accurate contact and temporal context for tracking stage progress.

In response, recent works augment VLAs with spatial cues or temporal context between the VL backbone and the action expert (Fig.~\ref{fig:teaser}(a), right), yet both dimensions are typically studied separately and exhibit inherent limitations when applied individually.
From a spatial perspective, existing methods incorporate 3D point and depth priors~\cite{qu2025spatialvla,li2025spatialforcing,zhang2025falcon,sun2026rocket} or spatial grounding and affordance cues~\cite{ye2026st4vla,liu2026palm} to enhance spatial awareness.
However, they typically model these priors in isolation without local contact geometry for gripper alignment, while those relying on test-time 3D encoders further incur inference latency.
From a temporal perspective, history-aware policies leverage multi-frame contexts, perceptual memories, or past motion cues~\cite{shi2025memoryvla,li2026cronusvla,koo2026hamlet,liu2026streampi,sridhar2026memer,yuan2026motionvla}, but remain confined to past observations without foresight for upcoming contacts and mostly rely on 2D historical features rather than explicit 3D contact geometry.
To provide such lookahead, world-model VLAs either generate future visual frames~\cite{tian2024seer,kim2026cosmospolicy,jin2026structvla} or predict latent visual and motion representations~\cite{zheng2025flare,lin2026hifvla,yang2026cowvla,wang2026mowam}.
Nevertheless, frame-level generation spends substantial computation on visual synthesis and incurs high inference latency, whereas 2D latent or motion predictions still lack fine-grained 3D contact cues for precise manipulation.

These limitations highlight that spatial perception and temporal reasoning are inherently coupled in robotic manipulation: each action relies on 3D physical cues for accurate contact, as well as past memory and future foresight to coordinate transitions across evolving 3D states.
On one hand, reliable contact control requires a unified 3D representation that jointly captures global scene geometry, task-relevant interaction regions, and local approach cues rather than isolated spatial priors.
On the other hand, evolving past, present, and future states directly over compact 3D tokens, rather than raw RGB pixels, provides explicit physical grounding across time steps while avoiding redundant visual computation and high inference latency.

Motivated by this insight, we propose \textbf{PhysPPF}, a VLA framework that performs unified spatiotemporal modeling in a compact physical token space (Fig.~\ref{fig:teaser}(b)).
\textbf{For spatial modeling}, we introduce \textbf{WWH} (\emph{What-Where-How}) to factorize 3D perception into three compact physical token groups: scene-level point and depth geometry capturing \emph{what} 3D structures occupy the workspace, instruction-conditioned affordances localizing \emph{where} to interact, and metric depth and local surface normals constraining \emph{how} the gripper should approach.
These tokens are supervised by pretrained perception teachers via lightweight decoders only during training, with both teachers and decoders discarded at inference to avoid test-time 3D encoding overhead.
\textbf{For temporal modeling}, we develop \textbf{PPF} (\emph{Past-Present-Future}) to model causal state transitions directly in this WWH space without pixel reconstruction.
Specifically, PPF consolidates post-execution states into a bounded \emph{past} physical memory via similarity-based merging, and employs a spatiotemporal world model conditioned on this memory and VL features to jointly estimate \emph{present} and predict \emph{future} physical tokens under asymmetric attention.
Finally, parallel gated residual cross-attention branches inject past, present, and future tokens into a flow-matching action expert to avoid cross-temporal interference, achieving strong manipulation performance with modest latency and parameter overhead (Fig.~\ref{fig:teaser}(c, d)).
Our contributions are as follows:

\begin{itemize}
    \item We propose \textbf{PhysPPF}, a VLA framework that performs unified spatiotemporal modeling in a compact physical token space. By integrating 3D spatial perception and temporal state evolution, it guides action generation across contact-rich and long-horizon tasks.
    
    \item We introduce \textbf{WWH} to factorize 3D workspace structure, task affordances, and local approach cues into compact physical tokens via training-only teacher supervision. In this space, \textbf{PPF} maintains a bounded past memory, infers present and future states via an asymmetric world model, and conditions actions through parallel gated branches.

    \item Extensive experiments across SimplerEnv, CALVIN, and LIBERO show that \textbf{PhysPPF} consistently outperforms existing spatial, history-aware, and world-model VLAs with modest latency and parameter overhead, improving over the StarVLA baseline by +17.2 points on SimplerEnv, +29.3 points in 5-subtask success on CALVIN, and +3.5 points on LIBERO.
\end{itemize}

\begin{figure*}[t]
    \centering
    \includegraphics[width=0.96\textwidth]{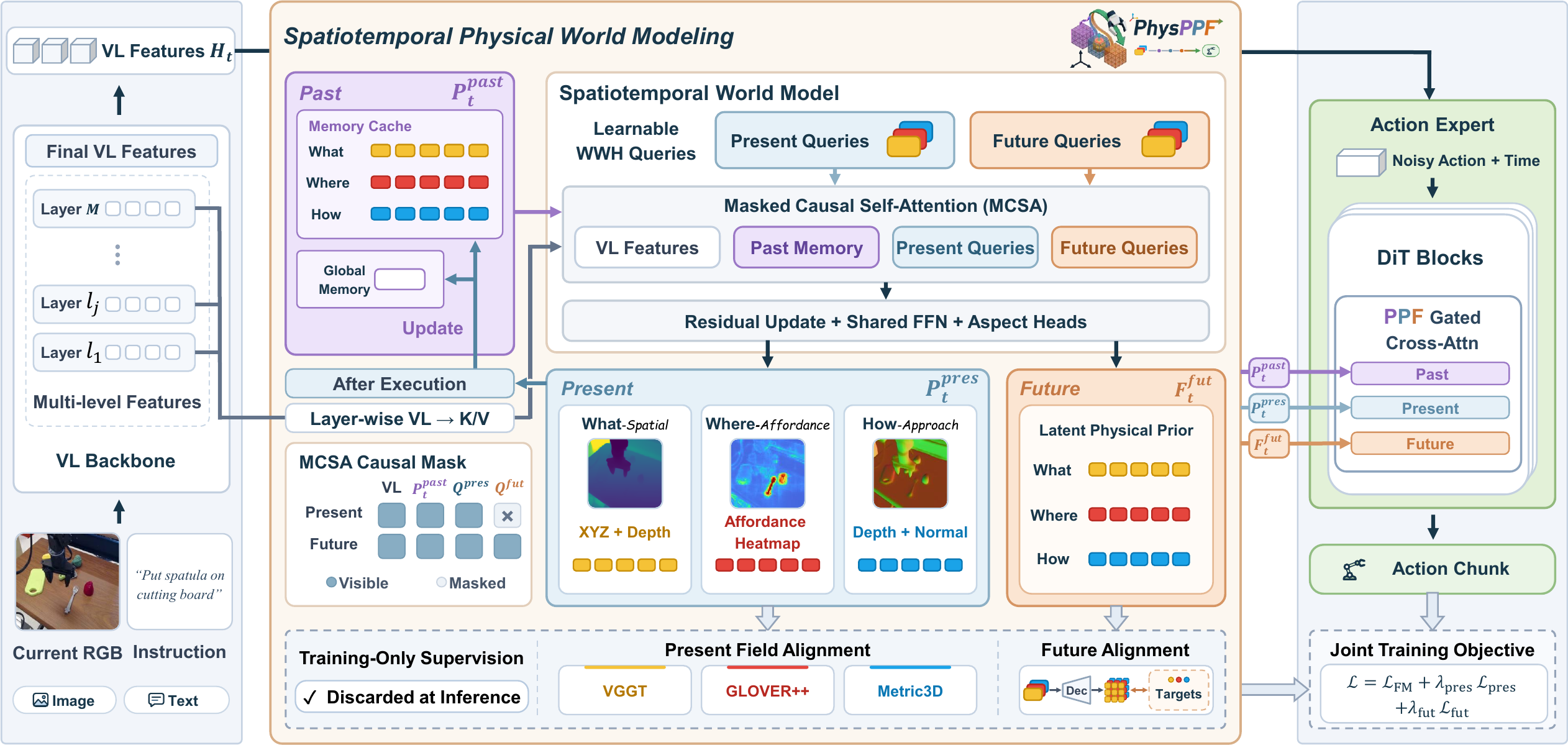}
    \caption{\textbf{Architecture overview.}
    \textbf{PhysPPF} is a spatiotemporal VLA framework that unifies 3D physical grounding, causal world modeling, and decoupled policy conditioning within a compact token space.
    A spatiotemporal world model combines multi-layer vision-language features with consolidated past memory ($P_t^{\mathit{past}}$) to jointly infer present ($P_t^{\mathit{pres}}$) and future ($F_t^{\mathit{fut}}$) \textbf{WWH} tokens under asymmetric attention.
    Each \textbf{PPF} state then conditions the flow-matching action expert through its own gated residual cross-attention branch.}
    \label{fig:overview}
\end{figure*}

% \begin{figure*}[t]
%     \centering
%     \includegraphics[width=0.96\textwidth]{assets/Fig2_v6.pdf}
%     % \caption{\textbf{Architecture overview.}
%     % \textbf{PhysPPF} is a spatiotemporal modeling VLA framework that unifies 3D physical grounding, causal world modeling, and decoupled policy conditioning within a compact token space.
%     % At its core, a unified physical Transformer leverages multi-layer vision-language features and consolidated past memory ($P_t$) to jointly infer present ($C_t$) and future ($F_t$) \textbf{WWH} tokens under asymmetric attention.
%     % Together, these \textbf{PPF} states independently guide a flow-matching action expert via branch-specific gated residual cross-attention.}
%     \caption{\textbf{Architecture overview.}
%     \textbf{PhysPPF} is a spatiotemporal VLA framework that unifies 3D physical grounding, causal world modeling, and decoupled policy conditioning within a compact token space.
%     A spatiotemporal world model combines multi-layer vision-language features with consolidated past memory ($P_t^{\mathit{past}}$) to jointly infer present ($P_t^{\mathit{pres}}$) and future ($F_t^{\mathit{fut}}$) \textbf{WWH} tokens under asymmetric attention.
%     Each \textbf{PPF} state then conditions the flow-matching action expert through its own gated residual cross-attention branch.}
%     \label{fig:overview}
% \end{figure*}

\section{Related Work}
\label{sec:related_work}
\noindent\textbf{Spatial Perception in VLA Models.}
Recent VLA models encode visual observations and language instructions with pretrained VL backbones to guide action generation~\cite{brohan2023rt2,kim2024openvla,black2024pi0,black2025pi05,nvidia2025grootn1,kim2025openvlaoft,community2026starvla}.
Although some studies refine visual perception via stronger vision encoders or active attention~\cite{zhang2026vlm4vla,xiao2026avavla}, they still rely on 2D image features without explicit 3D cues for contact-rich manipulation.
To enhance spatial awareness, one line of work incorporates 3D geometric priors, either injecting 3D positional and spatial tokens at test time~\cite{qu2025spatialvla,zhang2025falcon} or aligning VL features with 3D foundation models during training~\cite{li2025spatialforcing,sun2026rocket}.
Another line introduces spatial grounding and affordance cues via localization-guided training~\cite{ye2026st4vla} or progress-aware affordance prediction~\cite{liu2026palm}.
However, these methods mostly model scene geometry or 2D affordances separately without local surface cues for gripper alignment, while those using test-time 3D encoders add latency.
Moreover, most spatial VLAs are confined to single-frame perception rather than multi-step 3D evolution.
In contrast, \textbf{WWH} jointly encodes workspace geometry (\emph{what}), task affordances (\emph{where}), and local approach cues (\emph{how}) in a shared token space via training-only distillation, avoiding test-time 3D encoders while supporting multi-step physical modeling.

\noindent\textbf{Temporal and World Modeling in VLA Models.}
Beyond single-frame perception, long-horizon manipulation benefits from both retaining history to track stage progress and predicting future states to guide upcoming interactions.
History-aware VLAs capture past context via multi-frame inputs, streaming tokens, or motion cues~\cite{li2026cronusvla,koo2026hamlet,liu2026streampi,yuan2026motionvla}, while retrieval-based policies query perceptual tokens or experiences~\cite{shi2025memoryvla,sridhar2026memer}.
In parallel, world models forecast future dynamics by synthesizing dense or structured frames~\cite{tian2024seer,kim2026cosmospolicy,jin2026structvla}, or by aligning and predicting 2D visual latents and motion cues~\cite{zheng2025flare,lin2026hifvla,yang2026cowvla,bai2026bridgewa,wang2026mowam}.
Recent studies further model 3D spatiotemporal dynamics via calibrated RGB-D pretraining~\cite{zhang2025fourdivla}, training-only 3D flow~\cite{go2026flowvla}, or depth-lifted 3D motion~\cite{xiong2026geomvla}.
However, most methods model past memory and future prediction separately, lacking either contact foresight or consolidated long-horizon memory.
Moreover, frame generation incurs high latency, most history and latent methods operate in 2D spaces without explicit 3D contact cues, and recent 3D models either require calibrated depth inputs or lack test-time 3D foresight.
To bridge this gap, \textbf{PPF} evolves states directly in the compact WWH space from RGB inputs without pixel reconstruction, unifying bounded retrieval-free memory, present perception, and future prediction via an asymmetric world model and parallel gated branches.

\section{Methodology}
\label{sec:method}
\noindent\textbf{Overview.}
Fig.~\ref{fig:overview} illustrates the architecture of \textbf{PhysPPF}, which integrates a pretrained vision-language (VL) backbone, a spatiotemporal world model $\mathcal{U}_\phi$, and a flow-matching action expert $\mathcal{A}_\theta$.
Given the current RGB observation $I_t$ and language instruction $\ell$ at time step $t$, the $M$-layer VL backbone first extracts final features $H_t$ alongside $J$ multi-level intermediate features $\{H_t^{(l_j)}\}_{j=1}^{J}$ sampled at stride $M/J$ ($l_j=j\cdot M/J$).
The $J$-block world model $\mathcal{U}_\phi$ then interacts one-to-one with $\{H_t^{(l_j)}\}_{j=1}^{J}$, progressively aggregating visual-semantic cues from shallow to deep layers to model manipulation across space and time.
\textbf{Spatially}, \textbf{WWH} (\emph{What-Where-How}) defines a unified 3D representation for each physical state using three complementary token groups: global scene geometry (\emph{what}), instruction-conditioned affordances (\emph{where}), and local approach cues (\emph{how}) (Sec.~\ref{sec:wwh}).
\textbf{Temporally}, \textbf{PPF} (\emph{Past-Present-Future}) models causal state evolution directly over this WWH representation without pixel reconstruction (Sec.~\ref{sec:ppf}).

Specifically, WWH states from completed steps are consolidated into a \emph{past} memory $P_t^{\mathit{past}}$.
Conditioned on $\{H_t^{(l_j)}\}_{j=1}^{J}$ and $P_t^{\mathit{past}}$, $\mathcal{U}_\phi$ updates learnable WWH queries $(Q^{\mathit{pres}},Q^{\mathit{fut}})$ under asymmetric causal attention to jointly infer the \emph{present} state $P_t^{\mathit{pres}}$ and predict the \emph{future} state $F_t^{\mathit{fut}}$.
\textbf{Finally}, conditioned on the final VL features $H_t$, the action expert $\mathcal{A}_\theta$ further incorporates $P_t^{\mathit{past}}$, $P_t^{\mathit{pres}}$, and $F_t^{\mathit{fut}}$ through three parallel gated cross-attention branches to predict an $n$-step action chunk $\hat{\mathbf{A}}_t$:
\begin{equation}
\begin{aligned}
(P_t^{\mathit{pres}},F_t^{\mathit{fut}})&=\mathcal{U}_\phi\!\left(Q^{\mathit{pres}},Q^{\mathit{fut}},
\{H_t^{(l_j)}\}_{j=1}^{J},P_t^{\mathit{past}}\right),\\
\hat{\mathbf{A}}_t&=\mathcal{A}_\theta\!\left(H_t,P_t^{\mathit{past}},P_t^{\mathit{pres}},F_t^{\mathit{fut}}\right).
\end{aligned}
\label{eq:physppf_overview}
\end{equation}
After executing the initial $n_{\mathrm{exec}}$ actions of $\hat{\mathbf{A}}_t$ ($1\le n_{\mathrm{exec}}\le n$), the executed state $P_t^{\mathit{pres}}$ is committed to memory to form $P_{t+1}^{\mathit{past}}$ for step $t+1$, while frozen teachers supervise $(P_t^{\mathit{pres}},F_t^{\mathit{fut}})$ only during training (Sec.~\ref{sec:supervision}).

% ========================================================================================
\subsection{Manipulation-Relevant 3D Representation}
\label{sec:wwh}
Contact-rich manipulation requires workspace geometry, task-relevant affordances, and local approach cues without test-time 3D encoding latency.
We factorize spatial perception into three coupled \textbf{WWH} token groups within $\mathcal{U}_\phi$.

\par\noindent\textbf{What-Where-How Representation.}
We organize spatial cues into three aspects $\mathcal{K}=\{\mathrm{what},\mathrm{where},\mathrm{how}\}$: \emph{what} captures scene-level 3D geometry and obstacles, \emph{where} localizes instruction-conditioned affordances, and \emph{how} encodes metric-scale depth and surface-normal cues for gripper approach.
Specifically, learnable WWH queries $Q=[Q_{\mathrm{what}};Q_{\mathrm{where}};Q_{\mathrm{how}}]\in\mathbb{R}^{N\times d}$ comprise typed aspect groups $Q_k\in\mathbb{R}^{N_k\times d}$ ($k\in\mathcal{K}$), where $N_k$ is the token count of aspect $k$, $N=\sum_{k\in\mathcal{K}}N_k$, $d$ is the hidden dimension of $\mathcal{U}_\phi$, and $[\cdot\,;\cdot]$ denotes concatenation.
Across the $J$ shared Transformer blocks of $\mathcal{U}_\phi$, all three query groups jointly attend to multi-level VL features and to one another, exchanging complementary cues to yield updated aspect states $\widetilde{Q}=[\widetilde{Q}_{\mathrm{what}};\widetilde{Q}_{\mathrm{where}};\widetilde{Q}_{\mathrm{how}}]\in\mathbb{R}^{N\times d}$ ($\widetilde{Q}_k\in\mathbb{R}^{N_k\times d}$).
After the final block, distinct aspect heads $h_k:\mathbb{R}^d\to\mathbb{R}^{d_p}$ project $\widetilde{Q}_k$ into $d_p$-dimensional physical tokens:
\begin{equation}
P_k=h_k\big(\widetilde{Q}_k\big)\in\mathbb{R}^{N_k\times d_p},\qquad k\in\mathcal{K},
\label{eq:wwh_representation}
\end{equation}
forming the compact WWH physical representation $P=[P_{\mathrm{what}};P_{\mathrm{where}};P_{\mathrm{how}}]\in\mathbb{R}^{N\times d_p}$.

\begin{figure*}[t]
    \centering
    \includegraphics[width=0.92\textwidth]{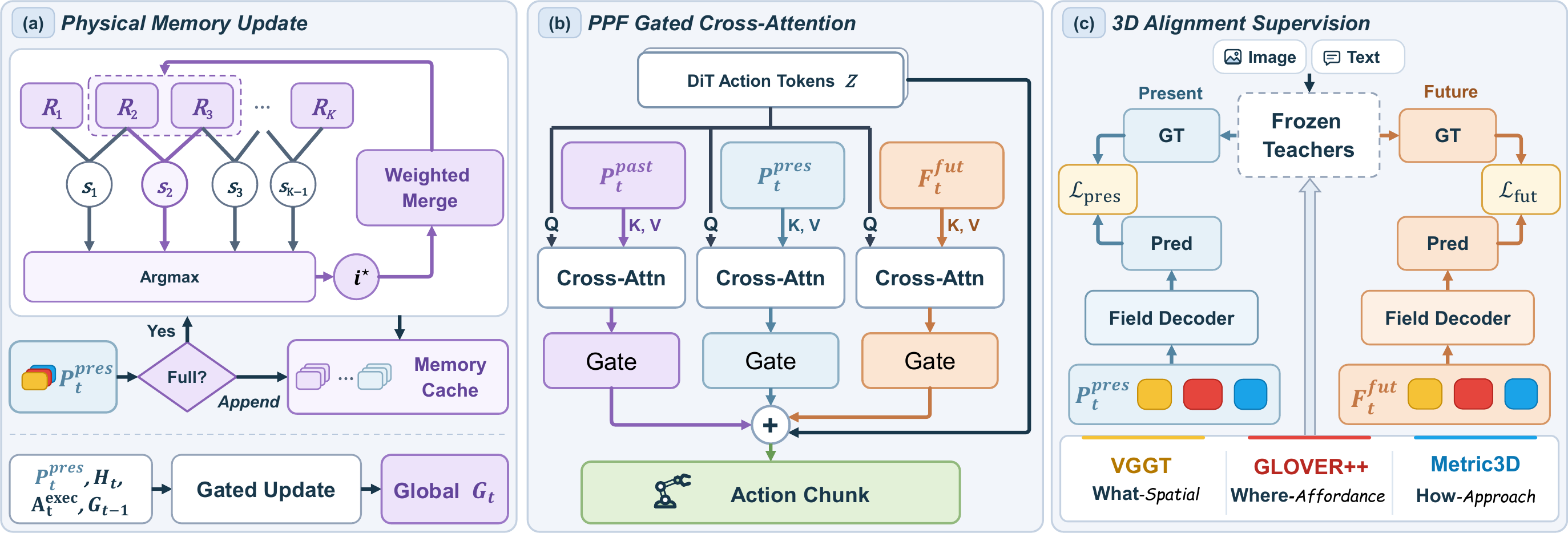}
    \caption{\textbf{Core module designs.}
    \textbf{(a) Physical Memory Update:} After execution, present WWH tokens ($P_t^{\mathit{pres}}$) enter a bounded local cache $\{R_i\}$ with similarity-based consolidation, while gated recurrence updates a global state $G_t$ to track task progress.
    \textbf{(b) PPF Gated Cross-Attention:} Parallel gated cross-attention branches condition DiT action tokens $Z$ on past ($P_t^{\mathit{past}}$), present ($P_t^{\mathit{pres}}$), and future ($F_t^{\mathit{fut}}$) physical contexts.
    \textbf{(c) 3D Alignment Supervision:} Training-only field decoders align present ($P_t^{\mathit{pres}}$) and future ($F_t^{\mathit{fut}}$) WWH tokens with geometric fields and affordance heatmaps from frozen VGGT, GLOVER++, and Metric3D teachers.}
    \label{fig:modules}
\end{figure*}

% ========================================================================================
\subsection{Temporal Physical World Modeling}
\label{sec:ppf}
Long-horizon manipulation requires tracking completed progress while anticipating upcoming physical transitions.
In the WWH space, \textbf{PPF} maintains a bounded past memory, jointly infers present and future states via asymmetric attention in $\mathcal{U}_\phi$, and guides $\mathcal{A}_\theta$ through parallel gated branches.

\par\noindent\textbf{Causal Physical Memory.}
We maintain a bounded dual-timescale memory (Fig.~\ref{fig:modules}(a)): a local cache $\mathcal{M}_{t-1}=\{R_i\}_{i=1}^{L_{t-1}}$ of capacity $K$ ($L_{t-1}\le K$) records executed WWH states $R_i=[R_{i,\mathrm{what}};R_{i,\mathrm{where}};R_{i,\mathrm{how}}]\in\mathbb{R}^{N\times d_p}$ ($R_{i,k}\in\mathbb{R}^{N_k\times d_p}$), each with a covered step count $c_i\ge 1$ (initialized to $1$), while a global state $G_{t-1}$ tracks overall task progress.
At step $t$, packing the cached states and projected global state $\Pi_G(G_{t-1})$ with token-type, record-time, and step-count embeddings yields retrieval-free past tokens:
\begin{equation}
(P_t^{\mathit{past}},M_t^{\mathit{past}})=\operatorname{Pack}\!\left(
\{R_i\}_{i=1}^{L_{t-1}},\Pi_G(G_{t-1})\right),
\label{eq:past_cache}
\end{equation}
where $M_t^{\mathit{past}}$ denotes the slot validity mask.
After executing the $n_{\mathrm{exec}}$-step action prefix $\mathbf{A}_t^{\mathrm{exec}}$ of $\hat{\mathbf{A}}_t$, we commit $P_t^{\mathit{pres}}\in\mathbb{R}^{N\times d_p}$ to $\mathcal{M}_t$ and update $G_t=\mathcal{R}(G_{t-1},P_t^{\mathit{pres}},H_t,\mathbf{A}_t^{\mathrm{exec}})$ via gated recurrence $\mathcal{R}$, excluding unexecuted action tails.
When the cache reaches capacity ($L_{t-1}=K$), we merge the most similar \emph{adjacent} pair before appending $P_t^{\mathit{pres}}$, preserving chronological order while compressing redundant quasi-static steps:
\begin{equation}
\begin{aligned}
s_i&=\sum_{k\in\mathcal{K}}
\operatorname{cos}(\bar{R}_{i,k},\bar{R}_{i+1,k}),\;
i^\star=\arg\max_{1\le i<K}s_i,\\
R_{\mathrm{merge}}&=
\frac{c_{i^\star}R_{i^\star}+c_{i^\star+1}R_{i^\star+1}}
{c_{i^\star}+c_{i^\star+1}},
\end{aligned}
\label{eq:memory_consolidation}
\end{equation}
where $\bar{R}_{i,k}\in\mathbb{R}^{d_p}$ is the mean-pooled token of $R_{i,k}$, $\operatorname{cos}(\cdot,\cdot)$ is cosine similarity, and $R_{\mathrm{merge}}$ replaces $(R_{i^\star},R_{i^\star+1})$ with count $c_{\mathrm{merge}}=c_{i^\star}+c_{i^\star+1}$.

\par\noindent\textbf{Asymmetric Present-Future World Modeling.}
Building on the WWH structure, we instantiate learnable present queries $Q^{\mathit{pres}}=[Q_{\mathrm{what}}^{\mathit{pres}};Q_{\mathrm{where}}^{\mathit{pres}};Q_{\mathrm{how}}^{\mathit{pres}}]$ and future queries $Q^{\mathit{fut}}=[Q_{\mathrm{what}}^{\mathit{fut}};Q_{\mathrm{where}}^{\mathit{fut}};Q_{\mathrm{how}}^{\mathit{fut}}]$ ($Q_k^{\mathit{pres}},Q_k^{\mathit{fut}}\in\mathbb{R}^{N_k\times d}$), initialized at step $t$ as $Q_t^{\mathit{pres},(0)}=Q^{\mathit{pres}}$ and $Q_t^{\mathit{fut},(0)}=Q^{\mathit{fut}}$.
Both streams share the $J$ blocks $\{\mathcal{T}_j\}_{j=1}^{J}$ of $\mathcal{U}_\phi$ under \emph{masked causal self-attention (MCSA)} (Fig.~\ref{fig:overview}).
Under MCSA, $Q^{\mathit{pres}}$ attends only to current VL features, past memory, and itself, whereas $Q^{\mathit{fut}}$ additionally attends unidirectionally to $Q^{\mathit{pres}}$, ensuring that present estimation remains strictly grounded in current observations and past memory without speculative leakage while future prediction builds upon the evolving present 3D state.
In block $j\in\{1,\ldots,J\}$, given projected VL features $\widetilde{H}_t^{(l_j)}$ and past tokens $\widetilde{P}_t^{\mathit{past}}$ (masked by $M_t^{\mathit{past}}$), $\mathcal{T}_j$ updates each stream $u\in\{\mathit{pres},\mathit{fut}\}$ over asymmetric contexts $\mathcal{S}_t^{u,(j)}$:
\begin{equation}
\begin{aligned}
\mathcal{S}_t^{\mathit{pres},(j)}
&=\big[\widetilde{H}_t^{(l_j)};\widetilde{P}_t^{\mathit{past}};Q_t^{\mathit{pres},(j-1)}\big],\\
\mathcal{S}_t^{\mathit{fut},(j)}
&=\big[\widetilde{H}_t^{(l_j)};\widetilde{P}_t^{\mathit{past}};
Q_t^{\mathit{pres},(j-1)};Q_t^{\mathit{fut},(j-1)}\big],\\
Q_t^{u,(j)}
&=\mathcal{T}_j\!\left(Q_t^{u,(j-1)};\mathcal{S}_t^{u,(j)}\right).
\end{aligned}
\label{eq:future_attention}
\end{equation}
After block $J$, stream-specific aspect heads $h_k^{\mathit{pres}},h_k^{\mathit{fut}}:\mathbb{R}^d\to\mathbb{R}^{d_p}$ project $Q_{t,k}^{\mathit{pres},(J)}$ and $Q_{t,k}^{\mathit{fut},(J)}$ ($k\in\mathcal{K}$):
\begin{equation}
P_{t,k}^{\mathit{pres}}=h_k^{\mathit{pres}}\big(Q_{t,k}^{\mathit{pres},(J)}\big),\;\;
F_{t,k}^{\mathit{fut}}=h_k^{\mathit{fut}}\big(Q_{t,k}^{\mathit{fut},(J)}\big),
\label{eq:ppf_projection}
\end{equation}
with $P_{t,k}^{\mathit{pres}},F_{t,k}^{\mathit{fut}}\in\mathbb{R}^{N_k\times d_p}$, forming both the inferred present state $P_t^{\mathit{pres}}=[P_{t,\mathrm{what}}^{\mathit{pres}};P_{t,\mathrm{where}}^{\mathit{pres}};P_{t,\mathrm{how}}^{\mathit{pres}}]\in\mathbb{R}^{N\times d_p}$ and predicted future state $F_t^{\mathit{fut}}=[F_{t,\mathrm{what}}^{\mathit{fut}};F_{t,\mathrm{where}}^{\mathit{fut}};F_{t,\mathrm{how}}^{\mathit{fut}}]\in\mathbb{R}^{N\times d_p}$.

\par\noindent\textbf{Parallel Temporal Conditioning.}
Because $P_t^{\mathit{past}}$ (subtask progress), $P_t^{\mathit{pres}}$ (current physical constraints), and $F_t^{\mathit{fut}}$ (lookahead cues) would compete under one attention softmax if concatenated, we introduce three \emph{parallel gated residual cross-attention branches} in the Diffusion Transformer (DiT) action expert $\mathcal{A}_\theta$ (Fig.~\ref{fig:modules}(b)).
Let $B_t^{r\in\{\mathit{past},\mathit{pres},\mathit{fut}\}}=(P_t^{\mathit{past}},P_t^{\mathit{pres}},F_t^{\mathit{fut}})$ denote the three temporal contexts.
At flow step $\tau\in[0,1]$, selected DiT blocks $b$ update noisy action tokens $Z_{t,\tau}^{(b)}$ into $\widetilde{Z}_{t,\tau}^{(b)}$:
\begin{equation}
\widetilde{Z}_{t,\tau}^{(b)}
=Z_{t,\tau}^{(b)}+
\sum_{r}g_{b,r}\,\operatorname{CrossAttn}_{b,r}\!\left(
Z_{t,\tau}^{(b)},B_t^r\right),
\label{eq:physical_conditioning}
\end{equation}
where $\operatorname{CrossAttn}_{b,r}$ queries $B_t^r$ from $Z_{t,\tau}^{(b)}$, $g_{b,r}$ is a learnable scalar gate balancing the three temporal contexts, and the $\mathit{past}$ branch applies validity mask $M_t^{\mathit{past}}$.

% ========================================================================================

\begin{table*}[t]
\centering
\caption{\textbf{Performance comparison on SimplerEnv (WidowX).}
Success rates (\%) on four manipulation tasks and their average.
Best and second-best results in each column are in \textbf{bold} and \underline{underlined}, respectively. Size indicates the backbone parameter count.}
\label{tab:simplerenv_main}
\begingroup
\footnotesize
\setlength{\tabcolsep}{4.5pt}
\renewcommand{\arraystretch}{1.06}
\begin{tabularx}{\textwidth}{
    lcc*{5}{>{\centering\arraybackslash}X}
}
\toprule
Method & Venue & Size
& \shortstack{Put\\Spoon}
& \shortstack{Put\\Carrot}
& \shortstack{Stack\\Block}
& \shortstack{Put\\Eggplant}
& Avg. $\uparrow$ \\
\midrule
$\pi_0$~\cite{black2024pi0} & RSS'25 & 3B
& 29.1 & 0.0 & 16.6 & 62.5 & 27.1 \\
SpatialVLA~\cite{qu2025spatialvla} & RSS'25 & 3B
& 16.7 & 25.0 & 29.2 & \textbf{100.0} & 42.7 \\
$\pi_{0.5}$~\cite{black2025pi05} & CoRL'25 & 3B
& 49.3 & \underline{64.7} & 44.7 & 69.7 & 57.1 \\
VLM4VLA~\cite{zhang2026vlm4vla} & ICLR'26 & 3B
& 79.2 & 20.8 & 0.0 & \underline{91.7} & 48.0 \\
MG-Select~\cite{jang2026mgselect} & ICLR'26 & 3B
& 69.4 & \textbf{75.0} & 43.1 & 13.9 & 50.3 \\
\midrule
OpenVLA-OFT~\cite{kim2025openvlaoft} & RSS'25 & 7B
& 12.5 & 4.2 & 4.2 & \textbf{100.0} & 30.2 \\
SP-VLA~\cite{li2026spvla} & ICLR'26 & 7B
& 70.8 & 54.2 & 29.2 & 75.0 & 57.3 \\
CronusVLA~\cite{li2026cronusvla} & AAAI'26 & 7B
& 66.7 & 54.2 & 20.8 & \textbf{100.0} & 60.4 \\
HAMLET~\cite{koo2026hamlet} & ICLR'26 & 7B
& 75.0 & 62.5 & 16.7 & \textbf{100.0} & 63.5 \\
MemoryVLA~\cite{shi2025memoryvla} & ICLR'26 & 7B
& 75.0 & \textbf{75.0} & 37.5 & \textbf{100.0} & 71.9 \\
StructVLA~\cite{jin2026structvla} & ECCV'26 & 8.5B
& 87.5 & \textbf{75.0} & \underline{45.8} & \underline{91.7} & \underline{75.0} \\
\midrule
Baseline (StarVLA)~\cite{community2026starvla}
& --- & 3B
& \underline{91.7} & 59.4 & 36.5 & 62.5 & 62.5 \\
\rowcolor[gray]{0.91}
\textbf{PhysPPF (Ours)} & --- & 3B
& \textbf{93.8} & 62.5 & \textbf{75.0} & 87.5 & \textbf{79.7} \\
\bottomrule
\end{tabularx}
\endgroup
\end{table*}

\begin{table*}[t]
\centering
\caption{\textbf{Performance comparison on CALVIN ABC$\rightarrow$D.} Success rates (\%) for completing at least $m$ consecutive subtasks ($m/5$) and the average completed sequence length (Avg.\ Len.). Best and second-best results in each column are in \textbf{bold} and \underline{underlined}, respectively.}
\label{tab:calvin_main}
\begingroup
\footnotesize
\setlength{\tabcolsep}{4.5pt}
\renewcommand{\arraystretch}{1.06}
\begin{tabularx}{\textwidth}{
    lcc*{6}{>{\centering\arraybackslash}X}
}
\toprule
Method & Venue & Size
& \multicolumn{5}{c}{Consecutive subtask success rate (\%) $\uparrow$}
& Avg. Len. $\uparrow$ \\
\cmidrule(lr){4-8}
& & & 1/5 & 2/5 & 3/5 & 4/5 & 5/5 & \\
\midrule
$\pi_0$~\cite{black2024pi0} & RSS'25 & 3B
& 93.8 & 85.0 & 76.7 & 68.1 & 59.9 & 3.84 \\
$\pi_{0.5}$~\cite{black2025pi05} & CoRL'25 & 3B
& 94.4 & 88.4 & 85.3 & 80.1 & 76.1 & 4.24 \\
MG-Select~\cite{jang2026mgselect} & ICLR'26 & 3B
& 96.9 & 88.0 & 77.8 & 67.6 & 55.8 & 3.86 \\
VLM4VLA~\cite{zhang2026vlm4vla} & ICLR'26 & 3B
& 92.2 & 84.2 & 76.6 & 70.0 & 62.6 & 3.86 \\
AtomicVLA~\cite{zhang2026atomicvla} & CVPR'26 & 3B
& 94.1 & 88.7 & 85.2 & 81.7 & 77.6 & 4.27 \\
StreamPI~\cite{liu2026streampi} & NeurIPS'26 & 3B
& 96.9 & 93.6 & 90.7 & 88.5 & \underline{85.0} & 4.55 \\
\midrule
OpenVLA-OFT~\cite{kim2025openvlaoft} & RSS'25 & 7B
& 96.3 & 89.1 & 82.4 & 75.8 & 66.5 & 4.10 \\
MemoryVLA~\cite{shi2025memoryvla} & ICLR'26 & 7B
& 94.8 & 87.4 & 81.4 & 75.9 & 69.4 & 4.09 \\
HiF-VLA~\cite{lin2026hifvla} & CVPR'26 & 7B
& \underline{98.5} & 94.1 & 88.1 & 81.4 & 73.1 & 4.35 \\
AVA-VLA~\cite{xiao2026avavla} & CVPR'26 & 7B
& \textbf{99.6} & \textbf{97.6} & \underline{94.1} & \underline{89.9} & 84.1 & \underline{4.65} \\
\midrule
Baseline (StarVLA)~\cite{community2026starvla}
& --- & 3B
& 91.7 & 81.9 & 72.7 & 65.3 & 58.1 & 3.70 \\
\rowcolor[gray]{0.91}
\textbf{PhysPPF (Ours)} & --- & 3B
& \underline{98.5} & \underline{97.0} & \textbf{94.5} & \textbf{91.4} & \textbf{87.4} & \textbf{4.69} \\
\bottomrule
\end{tabularx}
\endgroup
\end{table*}

\begin{table*}[t]
\centering
\caption{\textbf{Performance comparison on LIBERO.}
Success rates (\%) on the Spatial, Object, Goal, and Long task suites, and their average.
Best and second-best results in each column are in \textbf{bold} and \underline{underlined}, respectively.}
\label{tab:libero_main}
\begingroup
\footnotesize
\setlength{\tabcolsep}{4.5pt}
\renewcommand{\arraystretch}{1.06}
\begin{tabularx}{\textwidth}{
    lcc*{5}{>{\centering\arraybackslash}X}
}
\toprule
Method & Venue & Size
& Spatial & Object & Goal & Long & Avg. $\uparrow$ \\
\midrule
SpatialVLA~\cite{qu2025spatialvla} & RSS'25 & 3B
& 88.2 & 89.9 & 78.6 & 55.5 & 78.1 \\
$\pi_0$~\cite{black2024pi0} & RSS'25 & 3B
& 96.8 & 98.8 & 95.8 & 85.2 & 94.2 \\
AtomicVLA~\cite{zhang2026atomicvla} & CVPR'26 & 3B
& 98.8 & 98.8 & 97.2 & 96.2 & 97.8 \\
BehaviorVLA~\cite{hu2026behaviorvla} & ICML'26 & 3B
& \underline{99.2} & 99.4 & 98.8 & 94.6 & 98.0 \\
StreamPI~\cite{liu2026streampi} & NeurIPS'26 & 3B
& 98.8 & \textbf{99.8} & \textbf{99.6} & 95.0 & 98.3 \\
ACoT-VLA~\cite{zhong2026acotvla} & CVPR'26 & 3B
& 98.6 & 99.0 & \underline{99.4} & \underline{97.0} & \underline{98.5} \\
FlowVLA ($\pi_{0.5}$)~\cite{go2026flowvla} & CoRL'26 & 3B
& 99.0 & 98.8 & 97.4 & 94.8 & 97.5 \\
\midrule
OpenVLA-OFT~\cite{kim2025openvlaoft} & RSS'25 & 7B
& 97.7 & 98.0 & 96.1 & 95.3 & 96.8 \\
AVA-VLA~\cite{xiao2026avavla} & CVPR'26 & 7B
& 97.4 & 99.4 & 97.4 & \textbf{97.6} & 98.0 \\
Spatial Forcing~\cite{li2025spatialforcing} & ICLR'26 & 7B
& \textbf{99.4} & \underline{99.6} & 98.8 & 96.0 & \underline{98.5} \\
ROCKET~\cite{sun2026rocket} & NeurIPS'26 & 7B
& 98.2 & \textbf{99.8} & 98.8 & \underline{97.0} & \underline{98.5} \\
\midrule
Baseline (StarVLA)~\cite{community2026starvla}
& --- & 3B
& 97.8 & 98.2 & 94.6 & 90.8 & 95.4 \\
\rowcolor[gray]{0.91}
\textbf{PhysPPF (Ours)} & --- & 3B
& \textbf{99.4} & \underline{99.6} & \textbf{99.6} & 96.8 & \textbf{98.9} \\
\bottomrule
\end{tabularx}
\endgroup
\end{table*}

\subsection{Spatiotemporal Physical Supervision}
\label{sec:supervision}
To ground latent WWH states in physical priors across space and time without test-time latency, we supervise $(P_t^{\mathit{pres}},F_t^{\mathit{fut}})$ with frozen perception teachers via training-only field decoders alongside policy learning (Fig.~\ref{fig:modules}(c)).

\par\noindent\textbf{3D Alignment Supervision.}
At step $t$, frozen teachers extract dense present targets $\{Y_{t,k}^{\mathit{pres}}\}_{k\in\mathcal{K}}$ from $(I_t,\ell)$: VGGT~\cite{wang2025vggt} extracts global scene-consistent 3D point and relative depth maps $Y_{t,\mathrm{what}}^{\mathit{pres}}$ from $I_t$ for obstacle layout; GLOVER++~\cite{ma2025gloverpp} generates instruction-conditioned 2D affordance heatmaps $Y_{t,\mathrm{where}}^{\mathit{pres}}$ from $(I_t,\ell)$; and Metric3D~\cite{hu2024metric3dv2} complements global structure with metric-scale depth and local surface normals $Y_{t,\mathrm{how}}^{\mathit{pres}}$ from $I_t$ for fine-grained contact alignment.
Applying the same teachers to the valid lookahead frame $\Delta$ raw frames ahead yields future targets $\{Y_{t,k}^{\mathit{fut}}\}_{k\in\mathcal{K}}$.
For each aspect $k\in\mathcal{K}$, lightweight coordinate-query field decoders $\mathcal{D}_k^{\mathit{pres}}$ and $\mathcal{D}_k^{\mathit{fut}}$ cross-attend from normalized 2D grid coordinates to $P_{t,k}^{\mathit{pres}}$ and $F_{t,k}^{\mathit{fut}}$ to reconstruct $\hat{Y}_{t,k}^{\mathit{pres}}=\mathcal{D}_k^{\mathit{pres}}(P_{t,k}^{\mathit{pres}})$ and $\hat{Y}_{t,k}^{\mathit{fut}}=\mathcal{D}_k^{\mathit{fut}}(F_{t,k}^{\mathit{fut}})$, supervised on valid regions via
\begin{equation}
\begin{aligned}
\mathcal{L}_{\mathrm{pres}}&=\sum_{k\in\mathcal{K}}\alpha_k\,\mathcal{L}_k\big(\hat{Y}_{t,k}^{\mathit{pres}},Y_{t,k}^{\mathit{pres}}\big),\\
\mathcal{L}_{\mathrm{fut}}&=\sum_{k\in\mathcal{K}}\alpha_k\,\mathcal{L}_k\big(\hat{Y}_{t,k}^{\mathit{fut}},Y_{t,k}^{\mathit{fut}}\big),
\end{aligned}
\label{eq:cur_alignment}
\end{equation}
where $\mathcal{L}_k$ denotes the field alignment loss for aspect $k$ and $\alpha_k$ is its weight.

\par\noindent\textbf{End-to-End Training Objective.}
Given target action chunk $\mathbf{A}_t$, $\mathcal{A}_\theta$ minimizes the conditional flow-matching loss $\mathcal{L}_{\mathrm{FM}}$ by regressing the velocity field over $\tau\in[0,1]$ conditioned on $(H_t,P_t^{\mathit{past}},P_t^{\mathit{pres}},F_t^{\mathit{fut}})$.
Leveraging spatiotemporal field alignment, we jointly optimize the policy and training-only decoders $\{\mathcal{D}_k^{\mathit{pres}},\mathcal{D}_k^{\mathit{fut}}\}_{k\in\mathcal{K}}$ on causally ordered trajectories via
\begin{equation}
\mathcal{L}=\mathcal{L}_{\mathrm{FM}}
+\lambda_{\mathrm{pres}}\mathcal{L}_{\mathrm{pres}}
+\lambda_{\mathrm{fut}}\mathcal{L}_{\mathrm{fut}},
\label{eq:training_objective}
\end{equation}
where $\lambda_{\mathrm{pres}}$ and $\lambda_{\mathrm{fut}}$ weight present and future alignment.
Teachers and field decoders are discarded at inference.

\section{Experiments}
\label{sec:experiments}

We evaluate PhysPPF across three benchmarks (Sec.~\ref{sec:results}), followed by WWH and PPF ablations (Sec.~\ref{sec:ablation}), efficiency analysis (Sec.~\ref{sec:efficiency}), and qualitative rollouts (Sec.~\ref{sec:qualitative}).

\subsection{Experimental Setup}
\label{sec:experimental_setup}

% We evaluate \textbf{PhysPPF} on contact-sensitive manipulation, long-horizon task execution, and multi-task generalization.

\noindent\textbf{Benchmarks.}
SimplerEnv~\cite{li2024simpler} evaluates WidowX manipulation across four contact-sensitive tasks, reporting task-wise and average success rates (Avg.\ SR).
CALVIN~\cite{mees2022calvin} evaluates long-horizon language-conditioned manipulation under the ABC$\rightarrow$D setting, reporting success rates for completing at least $m\in\{1,\ldots,5\}$ consecutive subtasks and average completed sequence length (Avg.\ Len.).
LIBERO~\cite{liu2023libero} assesses manipulation across the \emph{Spatial}, \emph{Object}, \emph{Goal}, and \emph{Long} task suites, reporting suite-wise and average success rates.

\noindent\textbf{Baseline and Compared Methods.}
We build PhysPPF upon StarVLA~\cite{community2026starvla}, which pairs Qwen2.5-VL-3B~\cite{bai2025qwen25vl} with a GR00T~\cite{nvidia2025grootn1} flow-matching action expert, and use it as our direct baseline.
We also compare PhysPPF against 20 recent 3B--8.5B generalist, spatial, history-aware, and world-model VLAs under standard evaluation protocols.

\noindent\textbf{Implementation Details.}
The spatiotemporal world model $\mathcal{U}_\phi$ is a $J=12$-block Transformer ($d=768$) initialized from DINOv2 ViT-B/14~\cite{oquab2024dinov2}, paired block-wise with VL layers $\{l_j\}_{j=1}^{12}=\{3,6,\ldots,36\}$.
Each WWH aspect $k\in\mathcal{K}$ uses $N_k=16$ learnable queries ($N=48$), with aspect-specific heads producing physical tokens ($d_p=256$).
The causal physical memory stores up to $K=16$ local slots, and $\mathcal{A}_\theta$ predicts action chunks of length $n=16$.
Separate training-only field decoders $\{\mathcal{D}_k^{\mathit{pres}},\mathcal{D}_k^{\mathit{fut}}\}_{k\in\mathcal{K}}$ reconstruct present and future WWH fields on $24\times24$ coordinate grids.
We set lookahead offset $\Delta=8$ raw frames, $\lambda_{\mathrm{pres}}=0.05$, and $\lambda_{\mathrm{fut}}=0.01$ (details in the supplement).

\subsection{Evaluation Results}
\label{sec:results}

Tabs.~\ref{tab:simplerenv_main}--\ref{tab:libero_main} show that PhysPPF consistently outperforms existing generalist, spatial, history-aware, and world-model VLAs across all three benchmarks.

\begin{figure}[t]
    \centering
    \includegraphics[width=\linewidth]{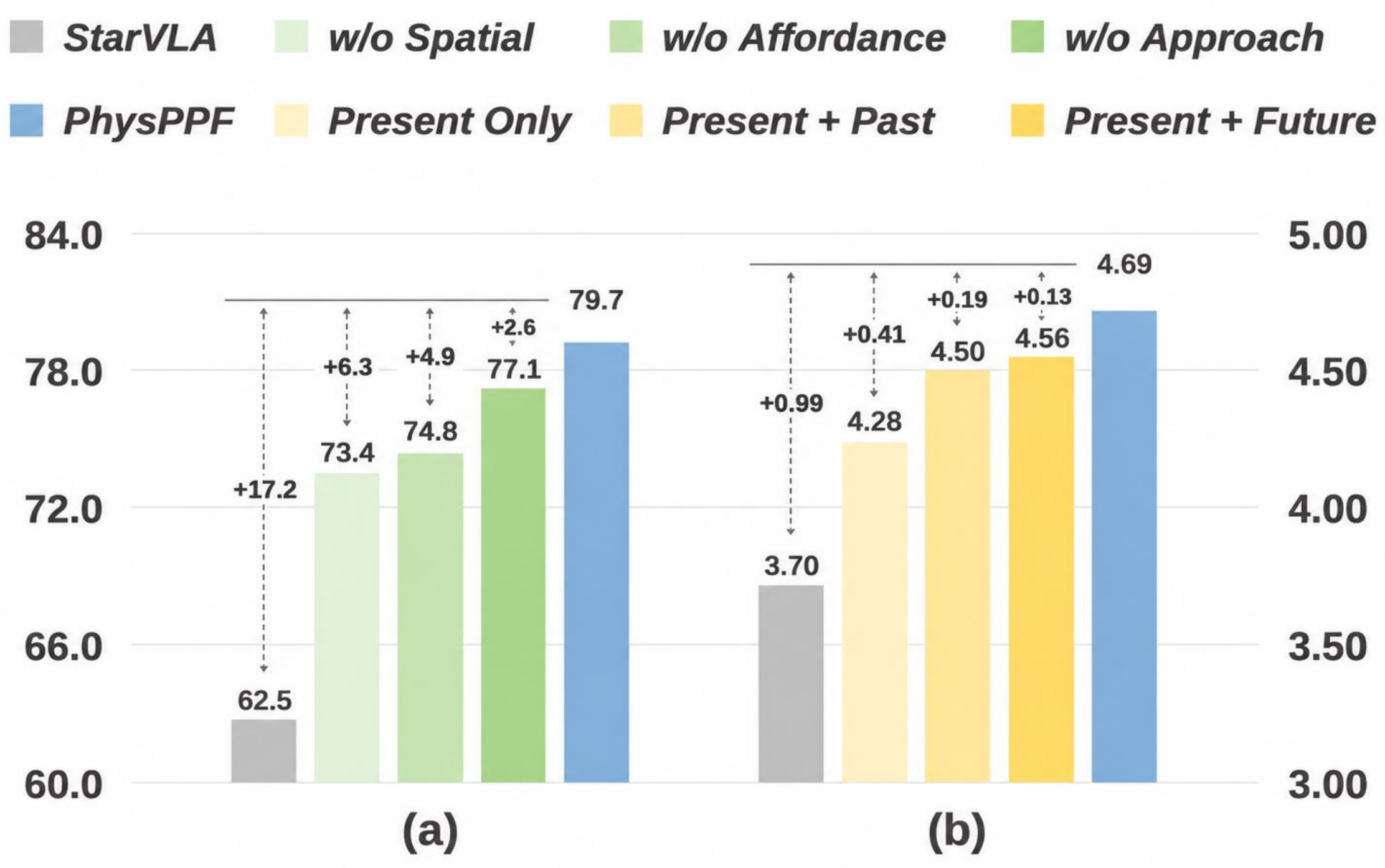}
    % \caption{\textbf{Ablation studies on WWH and PPF.}
    % \textbf{(a) WWH spatial ablation} on SimplerEnv (WidowX; Avg.\ SR, \%), individually removing \emph{what} (\emph{w/o Spatial}), \emph{where} (\emph{w/o Affordance}), or \emph{how} (\emph{w/o Approach}).
    % \textbf{(b) PPF temporal ablation} on CALVIN ABC$\rightarrow$D (Avg.\ Len.) across \emph{Present Only}, \emph{Present + Past}, \emph{Present + Future}, and \emph{Full PPF} (\textbf{PhysPPF}).}
    \caption{\textbf{Ablation studies on WWH and PPF.}
    (a) WWH spatial ablation on SimplerEnv (Avg.\ SR, \%), removing individual spatial cues (\emph{what}, \emph{where}, or \emph{how}).
    (b) PPF temporal ablation on CALVIN ABC$\rightarrow$D (Avg.\ Len.), evaluating different combinations of \emph{past}, \emph{present}, and \emph{future} representations.}
    \label{fig:ablation}
\end{figure}

\begin{figure*}[t]
    \centering
    \includegraphics[width=\textwidth]{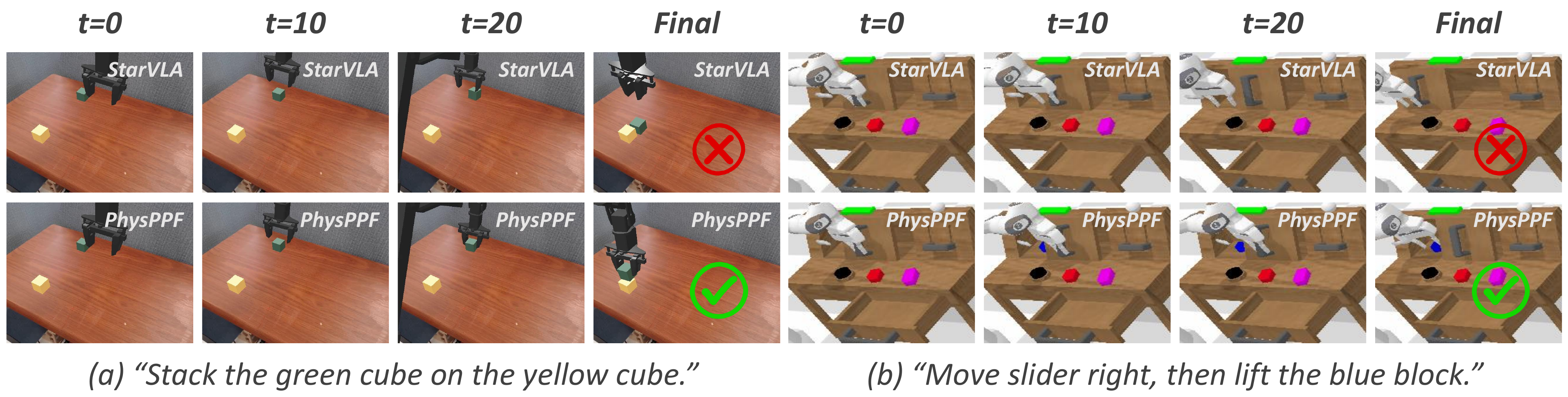}
    % \caption{\textbf{Qualitative comparison on SimplerEnv and CALVIN.}
    % Representative closed-loop rollouts of StarVLA (top) and \textbf{PhysPPF} (bottom) on
    % \textbf{(a)} SimplerEnv (WidowX) with the instruction \emph{``Stack the green cube on the yellow cube''} and
    % \textbf{(b)} CALVIN ABC$\rightarrow$D showing consecutive slider-moving and blue-block-lifting subtasks.}
    \caption{\textbf{Qualitative comparison on SimplerEnv and CALVIN.}
    Representative closed-loop rollouts of StarVLA (top) and PhysPPF (bottom) on (a) contact-rich block stacking in SimplerEnv and (b) sequential manipulation in CALVIN ABC$\rightarrow$D.}
    \label{fig:visualization}
\end{figure*}

\noindent\textbf{SimplerEnv.}
On SimplerEnv (Tab.~\ref{tab:simplerenv_main}), PhysPPF achieves 79.7\% Avg.\ SR (+17.2 points over baseline), surpassing 7B MemoryVLA~\cite{shi2025memoryvla} (71.9\%) and 8.5B StructVLA~\cite{jin2026structvla} (75.0\%) with a 3B VL backbone.
It improves all four tasks over baseline, raising \emph{Put Eggplant} to 87.5\% (+25.0 points) and more than doubling \emph{Stack Block} from 36.5\% to 75.0\% (+38.5 points).
Notably, all other methods reach at most 45.8\% on \emph{Stack Block}, revealing that our WWH and PPF effectively provide 3D clearance, gripper alignment, and state foresight for precise stacking.
% On SimplerEnv (Tab.~\ref{tab:simplerenv_main}), PhysPPF achieves 79.7\% Avg.\ SR (+17.2 points over StarVLA), surpassing 7B MemoryVLA~\cite{shi2025memoryvla} (71.9\%) and 8.5B StructVLA~\cite{jin2026structvla} (75.0\%) with only a 3B VL backbone.
% It improves all four tasks, raising \emph{Put Eggplant} to 87.5\% (+25.0 points) and more than doubling \emph{Stack Block} from 36.5\% to 75.0\% (+38.5 points).
% Notably, all other methods remain at or below 45.8\% on \emph{Stack Block}, revealing that our WWH spatial cues and future physical prediction provide the 3D clearance and gripper alignment critical for precise stacking.

\noindent\textbf{CALVIN.}
On CALVIN ABC$\rightarrow$D (Tab.~\ref{tab:calvin_main}), PhysPPF raises Avg.\ Len.\ from 3.70 to 4.69 (+26.8\% over baseline), outperforming 3B StreamPI~\cite{liu2026streampi} (4.55) and 7B AVA-VLA~\cite{xiao2026avavla} (4.65).
While all methods exceed 91\% on $1/5$, success rates degrade across chained subtasks, dropping from $1/5$ to $5/5$ by 33.6 points for StarVLA and 25.4 points for 7B MemoryVLA and HiF-VLA~\cite{lin2026hifvla}.
In contrast, PhysPPF limits this drop to 11.1 points and reaches 87.4\% on $5/5$ (+29.3 points over baseline), showing that past memory and future prediction support long-horizon execution.

\noindent\textbf{LIBERO.}
On LIBERO (Tab.~\ref{tab:libero_main}), PhysPPF reaches 98.9\% Avg.\ SR (+3.5 points over baseline), surpassing 3B ACoT-VLA~\cite{zhong2026acotvla}, 7B Spatial Forcing~\cite{li2025spatialforcing}, and 7B ROCKET~\cite{sun2026rocket} (all 98.5\%).
Beyond reaching 99.4\% on \emph{Spatial} and 99.6\% on \emph{Object}, it delivers larger gains on \emph{Goal} (99.6\%, +5.0 points) and \emph{Long} (96.8\%, +6.0 points), where diverse goals and multi-stage horizons benefit from instruction-conditioned affordances and temporal tracking.

% ========================================================================================
\subsection{Ablation Studies}
\label{sec:ablation}

% We evaluate \textbf{WWH} spatial components on SimplerEnv, where contact-sensitive manipulation emphasizes geometric precision, and \textbf{PPF} temporal components on CALVIN, where consecutive subtasks require long-horizon state tracking and coordination (Fig.~\ref{fig:ablation}).

As shown in Fig.~\ref{fig:ablation}, we ablate WWH spatial modeling on SimplerEnv, which demands precise 3D contact, and PPF temporal modeling on CALVIN ABC$\rightarrow$D, which requires long-horizon coordination across consecutive subtasks.

% \noindent\textbf{WWH Spatial Representation.}
% On SimplerEnv (Fig.~\ref{fig:ablation}(a)), removing global geometry (\emph{w/o Spatial}) decreases Avg.\ SR by 6.3 points, the largest drop among the three aspects.
% Removing instruction-conditioned affordances (\emph{w/o Affordance}) and approach cues (\emph{w/o Approach}) decreases Avg.\ SR by 4.9 and 2.6 points, respectively.
% All three partial variants remain at least 10.9 points above StarVLA, supporting the complementary roles of the WWH cues.

\noindent\textbf{WWH Spatial Modeling.}
On SimplerEnv (Fig.~\ref{fig:ablation}(a)), removing scene-level 3D geometry (\emph{w/o Spatial}) reduces Avg.\ SR from 79.7\% to 73.4\% ($-6.3$ points), the largest drop across aspects, as global 3D structure is foundational for spatial grounding.
Removing instruction-conditioned affordances (\emph{w/o Affordance}) and local approach cues (\emph{w/o Approach}) lowers Avg.\ SR to 74.8\% ($-4.9$ points) and 77.1\% ($-2.6$ points), reflecting less accurate contact localization and gripper alignment, respectively.
Nevertheless, all partial variants surpass StarVLA (62.5\%) by +10.9 to +14.6 points, confirming that \emph{what}, \emph{where}, and \emph{how} provide complementary spatial cues for contact-rich manipulation.

% \noindent\textbf{PPF Temporal World Modeling.}
% On CALVIN ABC$\rightarrow$D (Fig.~\ref{fig:ablation}(b)), \emph{Present Only} improves Avg.\ Len.\ from 3.70 to 4.28 (+0.58), demonstrating the benefit of current WWH grounding.
% Adding past memory (\emph{Present + Past}, 4.50) or future prediction (\emph{Present + Future}, 4.56) yields further gains of +0.22 and +0.28 over \emph{Present Only}, respectively.
% Combining all three contexts in \emph{Full PPF} reaches 4.69 (+0.19 and +0.13 over the two partial variants; +0.41 over \emph{Present Only}), supporting their complementary contributions.

\noindent\textbf{PPF Temporal Modeling.}
On CALVIN ABC$\rightarrow$D (Fig.~\ref{fig:ablation}(b)), conditioning solely on present WWH tokens already raises Avg.\ Len.\ from 3.70 to 4.28, as explicit 3D grounding reduces single-step manipulation errors.
Adding past memory or future prediction brings further gains of +0.22 and +0.28 by tracking completed progress and anticipating upcoming state transitions, respectively.
Combining all three contexts in PhysPPF reaches 4.69, surpassing \emph{Present + Past} by +0.19 and \emph{Present + Future} by +0.13, demonstrating that \emph{past}, \emph{present}, and \emph{future} provide complementary temporal context for long-horizon execution.

% ========================================================================================
\subsection{Efficiency Analysis}
\label{sec:efficiency}

% Tab.~\ref{tab:efficiency} compares active inference parameters, native per-chunk latency, and SimplerEnv success on an RTX 4090 GPU.
% Latency ratios reflect the tested native inference configurations and are not normalized by the number of executed actions.
Tab.~\ref{tab:efficiency} and Fig.~\ref{fig:teaser}(d) compare active inference parameters, native latency per policy call, and SimplerEnv success on an RTX 4090 GPU under each method's default settings.

\noindent\textbf{Spatial Efficiency.}
By discarding all perception teachers and field decoders at inference, PhysPPF adds only 0.208B active parameters and 18.7\,ms per policy call over baseline while improving Avg.\ SR by +17.2 points.
Compared with SpatialVLA~\cite{qu2025spatialvla}, which requires test-time 3D encoding, PhysPPF runs 3.80$\times$ faster (147.38 vs.\ 559.32\,ms) and achieves +37.0 points higher Avg.\ SR.

% \noindent\textbf{Temporal Efficiency.}
% By maintaining a bounded, retrieval-free memory and predicting future states in the compact WWH space without pixel reconstruction, PhysPPF uses nearly half the active parameters of retrieval-based MemoryVLA~\cite{shi2025memoryvla} and frame-generating StructVLA~\cite{jin2026structvla}.
% Meanwhile, it runs 1.76$\times$ and 8.62$\times$ faster per policy call while surpassing them by +7.8 and +4.7 points in Avg.\ SR.
\noindent\textbf{Temporal Efficiency.} By maintaining a bounded, retrieval-free memory and predicting future states in the compact WWH space without pixel reconstruction, PhysPPF enables a 3B-backbone policy to surpass 7B retrieval-based MemoryVLA~\cite{shi2025memoryvla} (+7.8 points) and 8.5B frame-generating StructVLA~\cite{jin2026structvla} (+4.7 points) with nearly half the active parameters, running 1.76$\times$ and 8.62$\times$ faster per policy call.

\begin{table}[t]
\centering
% \caption{\textbf{Performance-efficiency comparison on SimplerEnv (WidowX).}
% Active inference parameters, native per-chunk latency, and average success rate on an NVIDIA RTX 4090 GPU (BF16, batch size 1). Latencies follow each method's tested native inference configuration and are not normalized by executed actions. Best and second-best results in each column are in \textbf{bold} and \underline{underlined}, respectively.}
\caption{\textbf{Efficiency comparison on SimplerEnv (WidowX).}
Active parameters, native latency per policy call, and average success rate measured on an RTX 4090 GPU (BF16, batch size 1). Best and second-best results are in \textbf{bold} and \underline{underlined}.}
\label{tab:efficiency}
\begingroup
\footnotesize
\setlength{\tabcolsep}{2.5pt}
\renewcommand{\arraystretch}{1.08}
\begin{tabularx}{\columnwidth}{
    >{\raggedright\arraybackslash}Xccc
}
\toprule
Method
& Params (B) $\downarrow$
& Latency (ms) $\downarrow$
& Avg. SR (\%) $\uparrow$ \\
\midrule
SpatialVLA~\cite{qu2025spatialvla}
& \textbf{4.028} & 559.32 & 42.7 \\
MemoryVLA~\cite{shi2025memoryvla}
& 8.377 & 259.39 & 71.9 \\
StructVLA~\cite{jin2026structvla}
& 8.763 & 1269.93 & \underline{75.0} \\
\midrule
StarVLA~\cite{community2026starvla}
& \underline{4.320} & \textbf{128.67} & 62.5 \\
\rowcolor[gray]{0.91}
\textbf{PhysPPF (Ours)}
& 4.528 & \underline{147.38} & \textbf{79.7} \\
\bottomrule
\end{tabularx}
\endgroup
\end{table}

% ========================================================================================
\subsection{Qualitative Analysis}
\label{sec:qualitative}

% Fig.~\ref{fig:visualization} compares representative SimplerEnv and CALVIN rollouts of StarVLA and PhysPPF.
% \emph{Spatially}, in \emph{Stack Block} (Fig.~\ref{fig:visualization}(a)), StarVLA successfully grasps and transports the green cube ($T{=}0$--$20$) but releases it beside the yellow block (\emph{Final}), suggesting insufficient relative 3D grounding and local geometric awareness for precise placement.
% \emph{Temporally}, in CALVIN (Fig.~\ref{fig:visualization}(b)), the slider initially occludes the blue block.
% Although StarVLA begins moving the slider rightward ($T{=}10$--$20$), it reverses before sufficiently exposing the target (\emph{Final}), partially undoing its earlier progress.
% This behavior suggests difficulty retaining interaction history and anticipating the physical state needed to complete the subtask.
% In contrast, \textbf{PhysPPF} achieves more accurate placement and sustained task progression, consistent with the complementary benefits of \textbf{WWH} spatial grounding, causal physical memory, and future physical prediction.

Fig.~\ref{fig:visualization} compares representative closed-loop rollouts of StarVLA and PhysPPF.
\textbf{Spatially}, in \emph{Stack Block} (Fig.~\ref{fig:visualization}(a)), StarVLA approaches and grasps the green cube ($t{=}0$--$20$) but misaligns during transport and drops it beside the yellow cube (\emph{Final}), whereas PhysPPF leverages WWH spatial cues to accurately align the gripper and stack the cube on top.
\textbf{Temporally}, in CALVIN (Fig.~\ref{fig:visualization}(b)), the slider initially occludes the target blue block.
While StarVLA partially moves the slider rightward ($t{=}10$--$20$) before reversing and failing to expose the target (\emph{Final}), PhysPPF utilizes past memory and future prediction to fully open the slider ($t{=}10$--$20$) and transition to lifting the blue block (\emph{Final}).
\section{Conclusion}
\label{sec:conclusion}

% In this paper, we present \textbf{PhysPPF}, a VLA framework that performs unified spatiotemporal modeling in a compact physical token space for contact-rich and long-horizon manipulation.
% Spatially, \textbf{WWH} (\emph{What-Where-How}) distills pretrained perception teachers exclusively during training into compact token groups capturing scene-level 3D geometry (\emph{what}), instruction-conditioned affordances (\emph{where}), and local approach cues (\emph{how}), incurring no test-time 3D encoder overhead.
% Temporally, operating directly in this WWH space without pixel reconstruction, \textbf{PPF} (\emph{Past-Present-Future}) maintains a \emph{past} physical memory and jointly infers \emph{present} and \emph{future} physical tokens via an asymmetric world model, which then guide a flow-matching action expert through parallel gated branches.
% Extensive experiments across SimplerEnv, CALVIN, and LIBERO demonstrate that PhysPPF consistently outperforms existing generalist, spatial, history-aware, and world-model VLAs with modest parameter and latency overhead.

In this paper, we present \textbf{PhysPPF}, a VLA framework for unified spatiotemporal modeling in a compact physical token space. Spatially, \textbf{WWH} distills frozen teachers exclusively during training into compact tokens capturing scene geometry (\emph{what}), task affordances (\emph{where}), and local approach cues (\emph{how}), adding no test-time overhead. Temporally, operating in this WWH space without pixel reconstruction, \textbf{PPF} maintains a \emph{past} physical memory and jointly infers \emph{present} and \emph{future} tokens via an asymmetric world model to guide a flow-matching action expert through parallel gated branches. Experiments across SimplerEnv, CALVIN, and LIBERO show that PhysPPF consistently outperforms generalist, spatial, history-aware, and world-model VLAs with modest parameter and latency overhead.

\section*{LLM Usage Statement}
We used Large Language Models (specifically, OpenAI ChatGPT and Google Gemini) strictly as assistive tools to refine author-produced content, including polishing grammar and prose clarity, checking related-work phrasing and citation formatting, assisting with routine code debugging, and refining figure plotting scripts and visual presentation. No LLMs or generative AI tools were used for research ideation, hypothesis or claim formulation, data creation or annotation, or experimental design, execution, and result analysis. All AI-assisted edits were thoroughly reviewed and verified by the authors, who take full responsibility for all content in this manuscript.

{
    \small
    \bibliographystyle{ieeenat_fullname}
    \bibliography{main}

@String(CVPR  = {IEEE Conf. Comput. Vis. Pattern Recog.})

@String(ECCV  = {Eur. Conf. Comput. Vis.})

@String(AAAI  = {AAAI})

@inproceedings{zeng2021transporter,
  title         = {{Transporter Networks: Rearranging the Visual World for Robotic Manipulation}},
  author        = {Zeng, Andy and Florence, Pete and Tompson, Jonathan and Welker, Stefan and Chien, Jonathan and Attarian, Maria and Armstrong, Travis and Krasin, Ivan and Duong, Dan and Sindhwani, Vikas and others},
  booktitle     = {Proceedings of the 2020 Conference on Robot Learning},
  year          = {2021},
  volume        = {155},
  pages         = {726--747},
  month         = {16--18 Nov},
  editor        = {Kober, Jens and Ramos, Fabio and Tomlin, Claire},
  series        = {Proceedings of Machine Learning Research},
  publisher     = {PMLR},
  url           = {https://proceedings.mlr.press/v155/zeng21a.html},
}

@inproceedings{jang2022bcz,
  title         = {{BC-Z: Zero-Shot Task Generalization with Robotic Imitation Learning}},
  author        = {Jang, Eric and Irpan, Alex and Khansari, Mohi and Kappler, Daniel and Ebert, Frederik and Lynch, Corey and Levine, Sergey and Finn, Chelsea},
  booktitle     = {Proceedings of the 5th Conference on Robot Learning},
  year          = {2022},
  volume        = {164},
  pages         = {991--1002},
  month         = {08--11 Nov},
  editor        = {Faust, Aleksandra and Hsu, David and Neumann, Gerhard},
  series        = {Proceedings of Machine Learning Research},
  publisher     = {PMLR},
  url           = {https://proceedings.mlr.press/v164/jang22a.html},
}

@inproceedings{shridhar2022peract,
  title         = {{Perceiver-Actor: A Multi-Task Transformer for Robotic Manipulation}},
  author        = {Shridhar, Mohit and Manuelli, Lucas and Fox, Dieter},
  booktitle     = {Proceedings of The 6th Conference on Robot Learning},
  year          = {2023},
  volume        = {205},
  pages         = {785--799},
  month         = {14--18 Dec},
  editor        = {Liu, Karen and Kulic, Dana and Ichnowski, Jeff},
  series        = {Proceedings of Machine Learning Research},
  publisher     = {PMLR},
  url           = {https://proceedings.mlr.press/v205/shridhar23a.html},
}

@inproceedings{brohan2022rt1,
  title         = {{RT-1: Robotics Transformer for Real-World Control at Scale}},
  author        = {Anthony Brohan and Noah Brown and Justice Carbajal and Yevgen Chebotar and Joseph Dabis and Chelsea Finn and Keerthana Gopalakrishnan and Karol Hausman and Alexander Herzog and Jasmine Hsu and others},
  booktitle     = {Proceedings of Robotics: Science and Systems},
  year          = {2023},
  month         = {July},
  address       = {Daegu, Republic of Korea},
  doi           = {10.15607/RSS.2023.XIX.025},
}

@inproceedings{jiang2022vima,
  title         = {{{VIMA}: Robot Manipulation with Multimodal Prompts}},
  author        = {Jiang, Yunfan and Gupta, Agrim and Zhang, Zichen and Wang, Guanzhi and Dou, Yongqiang and Chen, Yanjun and Fei-Fei, Li and Anandkumar, Anima and Zhu, Yuke and Fan, Linxi},
  booktitle     = {Proceedings of the 40th International Conference on Machine Learning},
  year          = {2023},
  volume        = {202},
  pages         = {14975--15022},
  month         = {23--29 Jul},
  editor        = {Krause, Andreas and Brunskill, Emma and Cho, Kyunghyun and Engelhardt, Barbara and Sabato, Sivan and Scarlett, Jonathan},
  series        = {Proceedings of Machine Learning Research},
  publisher     = {PMLR},
  url           = {https://proceedings.mlr.press/v202/jiang23b.html},
}

@inproceedings{chi2023diffusionpolicy,
  title         = {{Diffusion Policy: Visuomotor Policy Learning via Action Diffusion}},
  author        = {Cheng Chi and Siyuan Feng and Yilun Du and Zhenjia Xu and Eric Cousineau and Benjamin CM Burchfiel and Shuran Song},
  booktitle     = {Proceedings of Robotics: Science and Systems},
  year          = {2023},
  month         = {July},
  address       = {Daegu, Republic of Korea},
  doi           = {10.15607/RSS.2023.XIX.026},
}

@inproceedings{zhao2023act,
  title         = {{Learning Fine-Grained Bimanual Manipulation with Low-Cost Hardware}},
  author        = {Tony Z. Zhao and Vikash Kumar and Sergey Levine and Chelsea Finn},
  booktitle     = {Proceedings of Robotics: Science and Systems},
  year          = {2023},
  month         = {July},
  address       = {Daegu, Republic of Korea},
  doi           = {10.15607/RSS.2023.XIX.016},
}

@inproceedings{openx2023rtx,
  title         = {{Open X-Embodiment: Robotic Learning Datasets and RT-X Models}},
  author        = {{Open X-Embodiment Collaboration}},
  booktitle     = {2024 IEEE International Conference on Robotics and Automation (ICRA)},
  year          = {2024},
  pages         = {6892--6903},
  doi           = {10.1109/ICRA57147.2024.10611477},
}

@inproceedings{octo2024,
  title         = {{Octo: An Open-Source Generalist Robot Policy}},
  author        = {Dibya Ghosh and Homer Rich Walke and Karl Pertsch and Kevin Black and Oier Mees and Sudeep Dasari and Joey Hejna and Tobias Kreiman and Charles Xu and Jianlan Luo and others},
  booktitle     = {Proceedings of Robotics: Science and Systems},
  year          = {2024},
  month         = {July},
  address       = {Delft, Netherlands},
  doi           = {10.15607/RSS.2024.XX.090},
}

@inproceedings{brohan2023rt2,
  title         = {{RT-2: Vision-Language-Action Models Transfer Web Knowledge to Robotic Control}},
  author        = {Zitkovich, Brianna and Yu, Tianhe and Xu, Sichun and Xu, Peng and Xiao, Ted and Xia, Fei and Wu, Jialin and Wohlhart, Paul and Welker, Stefan and Wahid, Ayzaan and others},
  booktitle     = {Proceedings of The 7th Conference on Robot Learning},
  year          = {2023},
  volume        = {229},
  pages         = {2165--2183},
  month         = {06--09 Nov},
  editor        = {Tan, Jie and Toussaint, Marc and Darvish, Kourosh},
  series        = {Proceedings of Machine Learning Research},
  publisher     = {PMLR},
  url           = {https://proceedings.mlr.press/v229/zitkovich23a.html},
}

@inproceedings{kim2024openvla,
  title         = {{OpenVLA: An Open-Source Vision-Language-Action Model}},
  author        = {Kim, Moo Jin and Pertsch, Karl and Karamcheti, Siddharth and Xiao, Ted and Balakrishna, Ashwin and Nair, Suraj and Rafailov, Rafael and Foster, Ethan P and Sanketi, Pannag R and Vuong, Quan and others},
  booktitle     = {Proceedings of The 8th Conference on Robot Learning},
  year          = {2025},
  volume        = {270},
  pages         = {2679--2713},
  month         = {06--09 Nov},
  editor        = {Agrawal, Pulkit and Kroemer, Oliver and Burgard, Wolfram},
  series        = {Proceedings of Machine Learning Research},
  publisher     = {PMLR},
  url           = {https://proceedings.mlr.press/v270/kim25c.html},
}

@inproceedings{black2024pi0,
  title         = {{$\pi_{0}$: A Vision-Language-Action Flow Model for General Robot Control}},
  author        = {Kevin Black and Noah Brown and Danny Driess and Adnan Esmail and Michael Robert Equi and Chelsea Finn and Niccolo Fusai and Lachy Groom and Karol Hausman and Brian Ichter and others},
  booktitle     = {Proceedings of Robotics: Science and Systems},
  year          = {2025},
  month         = {June},
  address       = {LosAngeles, CA, USA},
  doi           = {10.15607/RSS.2025.XXI.010},
}

@inproceedings{black2025pi05,
  title         = {{$\pi_{0.5}$: a Vision-Language-Action Model with Open-World Generalization}},
  author        = {Black, Kevin and Brown, Noah and Darpinian, James and Dhabalia, Karan and Driess, Danny and Esmail, Adnan and Equi, Michael Robert and Finn, Chelsea and Fusai, Niccolo and Galliker, Manuel Y. and others},
  booktitle     = {Proceedings of The 9th Conference on Robot Learning},
  year          = {2025},
  volume        = {305},
  pages         = {17--40},
  month         = {27--30 Sep},
  editor        = {Lim, Joseph and Song, Shuran and Park, Hae-Won},
  series        = {Proceedings of Machine Learning Research},
  publisher     = {PMLR},
  url           = {https://proceedings.mlr.press/v305/black25a.html},
}

@inproceedings{kim2025openvlaoft,
  title         = {{Fine-Tuning Vision-Language-Action Models: Optimizing Speed and Success}},
  author        = {Moo Jin Kim and Chelsea Finn and Percy Liang},
  booktitle     = {Proceedings of Robotics: Science and Systems},
  year          = {2025},
  month         = {June},
  address       = {LosAngeles, CA, USA},
  doi           = {10.15607/RSS.2025.XXI.017},
}

@article{nvidia2025grootn1,
  title         = {{GR00T N1: An Open Foundation Model for Generalist Humanoid Robots}},
  author        = {{NVIDIA} and Johan Bjorck and Fernando Casta{\~{n}}eda and Nikita Cherniadev and Xingye Da and Runyu Ding and Linxi ``Jim" Fan and Yu Fang and Dieter Fox and Fengyuan Hu and others},
  journal       = {arXiv preprint arXiv:2503.14734},
  year          = {2025},
  eprint        = {2503.14734},
  archivePrefix = {arXiv},
  primaryClass  = {cs.RO},
  url           = {https://arxiv.org/abs/2503.14734},
}

@inproceedings{qu2025spatialvla,
  title         = {{SpatialVLA: Exploring Spatial Representations for Visual-Language-Action Models}},
  author        = {Delin Qu and Haoming Song and Qizhi Chen and Yuanqi Yao and Xinyi Ye and Jiayuan Gu and Zhigang Wang and Yan Ding and Bin Zhao and Dong Wang and others},
  booktitle     = {Proceedings of Robotics: Science and Systems},
  year          = {2025},
  month         = {June},
  address       = {LosAngeles, CA, USA},
  doi           = {10.15607/RSS.2025.XXI.011},
}

@inproceedings{li2025spatialforcing,
  title         = {{Spatial Forcing: Implicit Spatial Representation Alignment for Vision-language-action Model}},
  author        = {Li, Fuhao and Song, Wenxuan and Zhao, Han and Wang, Jingbo and Ding, Pengxiang and Wang, Donglin and Zeng, Long and Li, Haoang},
  booktitle     = {International Conference on Learning Representations},
  year          = {2026},
  volume        = {2026},
  pages         = {132324--132345},
  editor        = {C. Vondrick and B. Hariharan and C. Raffel and L. Pinto and D. Yang and A. Faust},
  url           = {https://proceedings.iclr.cc/paper_files/paper/2026/file/d676c05a8fb93ad9231508d95567c498-Paper-Conference.pdf},
}

@inproceedings{ye2026st4vla,
  title         = {{Spatially Guided Training for Vision-Language-Action Model}},
  author        = {Jinhui Ye and Fangjing Wang and Ning Gao and Junqiu Yu and Zhu, Yangkun and Bin Wang and Jinyu Zhang and Weiyang Jin and Yanwei Fu and Feng Zheng and others},
  booktitle     = {The Fourteenth International Conference on Learning Representations},
  year          = {2026},
  url           = {https://openreview.net/forum?id=eKhOrQWAVJ},
}

@inproceedings{zhang2025falcon,
  title         = {{From Spatial to Actions: Grounding Vision-Language-Action Model in Spatial Foundation Priors}},
  author        = {Zhengshen Zhang and Hao Li and Yalun Dai and Zhengbang Zhu and Lei Zhou and Chenchen Liu and Dong Wang and Francis E. H. Tay and Sijin Chen and Ziwei Liu and others},
  booktitle     = {The Fourteenth International Conference on Learning Representations},
  year          = {2026},
  url           = {https://openreview.net/forum?id=fzmittHfq3},
}

@inproceedings{liu2026palm,
  title         = {{PALM: Progress-Aware Policy Learning via Affordance Reasoning for Long-Horizon Robotic Manipulation}},
  author        = {Liu, Yuanzhe and Zhu, Jingyuan and Mo, Yuchen and Li, Gen and Cao, Xu and Jin, Jin and Shen, Yifan and Li, Zhengyuan and Yu, Tianjiao and Yuan, Wenzhen and others},
  booktitle     = {Proceedings of the IEEE/CVF Conference on Computer Vision and Pattern Recognition (CVPR)},
  year          = {2026},
  pages         = {28096--28110},
  month         = {June},
}

@inproceedings{sun2026rocket,
  title         = {{ROCKET}: Residual-Oriented Multi-Layer Alignment for Spatially-Aware Vision-Language-Action Models},
  author        = {Sun, Guoheng and Du, Tingting and Feng, Kaixi and Luo, Chenxiang and Ding, Xingguo and Shen, Zheyu and Wang, Ziyao and He, Yexiao and Li, Ang},
  booktitle     = {Advances in Neural Information Processing Systems},
  year          = {2026},
  url           = {https://arxiv.org/abs/2602.17951},
}

@inproceedings{xiong2026geomvla,
  title         = {{{GeomVLA}: Unifying Scene, Motion, and Action in 3D}},
  author        = {Xiong, Ziyin and Gkanatsios, Nikos and Reuss, Moritz and Fragkiadaki, Katerina},
  booktitle     = {Proceedings of the Conference on Robot Learning (CoRL)},
  year          = {2026},
}

@inproceedings{shi2025memoryvla,
  title         = {{Memory{VLA}: Perceptual-Cognitive Memory in Vision-Language-Action Models for Robotic Manipulation}},
  author        = {Hao Shi and Bin Xie and Yingfei Liu and Lin Sun and Fengrong Liu and Tiancai Wang and Erjin Zhou and Haoqiang Fan and Xiangyu Zhang and Gao Huang},
  booktitle     = {The Fourteenth International Conference on Learning Representations},
  year          = {2026},
  url           = {https://openreview.net/forum?id=54U3XHf7qq},
}

@inproceedings{tian2024seer,
  title         = {{Predictive Inverse Dynamics Models are Scalable Learners for Robotic Manipulation}},
  author        = {Tian, Yang and Yang, Sizhe and Zeng, Jia and Wang, Ping and Lin, Dahua and Dong, Hao and Pang, Jiangmiao},
  booktitle     = {International Conference on Learning Representations},
  year          = {2025},
  volume        = {2025},
  pages         = {92033--92052},
  editor        = {Y. Yue and A. Garg and N. Peng and F. Sha and R. Yu},
  url           = {https://proceedings.iclr.cc/paper_files/paper/2025/file/e5b5c402bb7bd5e60bede6961d6fe39e-Paper-Conference.pdf},
}

@inproceedings{zheng2025flare,
  title         = {{FLARE: Robot Learning with Implicit World Modeling}},
  author        = {Zheng, Ruijie and Wang, Jing and Reed, Scott and Bjorck, Johan and Fang, Yu and Hu, Fengyuan and Jang, Joel and Kundalia, Kaushil and Lin, Zongyu and Magne, Lo\"{i}c and others},
  booktitle     = {Proceedings of The 9th Conference on Robot Learning},
  year          = {2025},
  volume        = {305},
  pages         = {3952--3971},
  month         = {27--30 Sep},
  editor        = {Lim, Joseph and Song, Shuran and Park, Hae-Won},
  series        = {Proceedings of Machine Learning Research},
  publisher     = {PMLR},
  url           = {https://proceedings.mlr.press/v305/zheng25a.html},
}

@article{bai2026bridgewa,
  title         = {{Bridge-WA: Learning Action-Relevant World Dynamics for Robotic Manipulation}},
  author        = {Bai, Yongjie and Dai, Mingtong and Wang, Zhouxia and Wang, Hanting and Zhong, Qijun and Yan, Feng and Liu, Yang and Lin, Liang},
  journal       = {arXiv preprint arXiv:2607.02195},
  year          = {2026},
  url           = {https://arxiv.org/abs/2607.02195},
}

@inproceedings{lin2026hifvla,
  title         = {{HiF-VLA: Hindsight, Insight and Foresight through Motion Representation for Vision-Language-Action Models}},
  author        = {Lin, Minghui and Ding, Pengxiang and Wang, Shu and Zhuang, Zifeng and Liu, Yang and Tong, Xinyang and Song, Wenxuan and Lyu, Shangke and Huang, Siteng and Wang, Donglin},
  booktitle     = {Proceedings of the IEEE/CVF Conference on Computer Vision and Pattern Recognition (CVPR)},
  year          = {2026},
  pages         = {20732--20742},
  month         = {June},
}

@article{li2026cronusvla,
  title         = {{Towards Efficient and Robust Manipulation via Multi-Frame Vision-Language-Action Modeling}},
  author        = {Li, Hao and Yang, Shuai and Chen, Yilun and Chen, Xinyi and Yang, Xiaoda and Tian, Yang and Wang, Hanqing and Wang, Tai and Lin, Dahua and Zhao, Feng and others},
  journal       = {Proceedings of the AAAI Conference on Artificial Intelligence},
  year          = {2026},
  volume        = {40},
  number        = {22},
  pages         = {18388--18396},
  month         = {Mar.},
  doi           = {10.1609/aaai.v40i22.38903},
  url           = {https://ojs.aaai.org/index.php/AAAI/article/view/38903},
  abstractNote  = {Recent vision-language-action (VLA) models built on pretrained vision-language models (VLMs) have demonstrated strong performance in robotic manipulation. However, these models remain constrained by the single-frame image paradigm and fail to fully leverage the temporal information offered by multi-frame histories, as directly feeding multiple frames into VLM backbones incurs substantial computational overhead and inference latency. We propose CronusVLA, a unified framework that extends single-frame VLA models to the multi-frame paradigm. CronusVLA follows a two-stage process: (1) Single-frame pretraining on large-scale embodied datasets with autoregressive prediction of action tokens, establishing an effective embodied vision-language foundation; (2) Multi-frame post-training, which adapts the prediction of the vision-language backbone from discrete tokens to learnable features, and aggregates historical information via feature chunking. CronusVLA effectively addresses the existing challenges of multi-frame modeling while enhancing performance. To evaluate the robustness under temporal and spatial disturbances, we introduce SimplerEnv-OR, a novel benchmark featuring 24 types of observational disturbances and 120 severity levels. Experiments across three embodiments in simulated and real-world environments demonstrate that CronusVLA achieves leading performance and superior robustness, with a 70.9% success rate on SimplerEnv, a 26.8% improvement over OpenVLA on LIBERO, and the highest robustness score on SimplerEnv-OR, showing the promise of efficient multi-frame adaptation for real-world VLA deployment.},
}

@inproceedings{yuan2026motionvla,
  title         = {{MotionVLA: Injecting Geometric Motion into Vision-Language-Action Model}},
  author        = {Yuan, Shanglin and Zhao, Weiheng and Guo, Xianda and Sui, Wei and Yu, Li and Liu, Wenyu and Wang, Xinggang},
  booktitle     = {Conference on Robot Learning (CoRL)},
  year          = {2026},
}

@inproceedings{sridhar2026memer,
  title         = {{Scaling up Memory for Robotic Control via Experience Retrieval}},
  author        = {Sridhar, Ajay and Pan, Jennifer and Sharma, Satvik and Finn, Chelsea},
  booktitle     = {International Conference on Learning Representations},
  year          = {2026},
  volume        = {2026},
  pages         = {97142--97166},
  editor        = {C. Vondrick and B. Hariharan and C. Raffel and L. Pinto and D. Yang and A. Faust},
  url           = {https://proceedings.iclr.cc/paper_files/paper/2026/file/9da515b1ad19d032a7398f00f5ff9b0c-Paper-Conference.pdf},
}

@article{wang2026mowam,
  title         = {{MoWAM: Explicit Future Motion Prediction for Efficient World Action Models}},
  author        = {Jiayu Wang and Bin Zhu and Yue Yu and Jingjing Chen},
  journal       = {arXiv preprint arXiv:2609.20709},
  year          = {2026},
  eprint        = {2609.20709},
  archivePrefix = {arXiv},
  primaryClass  = {cs.RO},
  url           = {https://arxiv.org/abs/2609.20709},
}

@inproceedings{zhang2025fourdivla,
  title         = {{4D-VLA: Spatiotemporal Vision-Language-Action Pretraining with Cross-Scene Calibration}},
  author        = {Zhang, Jiahui and Chen, Yurui and Xu, Yueming and Huang, Ze and Zhou, Yanpeng and Yuan, Yu-Jie and Cai, Xinyue and Huang, Guowei and Quan, Xingyue and Xu, Hang and others},
  booktitle     = {Advances in Neural Information Processing Systems},
  year          = {2025},
  volume        = {38},
  pages         = {33914--33937},
  editor        = {D. Belgrave and C. Zhang and H. Lin and R. Pascanu and P. Koniusz and M. Ghassemi and N. Chen},
  publisher     = {Curran Associates, Inc.},
  doi           = {10.52202/085713-1139},
  url           = {https://proceedings.neurips.cc/paper_files/paper/2025/file/30b9c38b9ebeee281cd2bc41d39bf0e7-Paper-Conference.pdf},
}

@inproceedings{xiao2026avavla,
  title         = {{AVA-VLA: Improving Vision-Language-Action models with Active Visual Attention}},
  author        = {Xiao, Lei and Li, Jifeng and Gao, Juntao and Ye, Feiyang and Jin, Yan and Qian, Jingjing and Zhang, Jing and Wu, Yong and Yu, Xiaoyuan},
  booktitle     = {Proceedings of the IEEE/CVF Conference on Computer Vision and Pattern Recognition (CVPR)},
  year          = {2026},
  pages         = {13453--13463},
  month         = {June},
}

@inproceedings{koo2026hamlet,
  title         = {{{HAMLET}: Switch Your Vision-Language-Action Model into a History-Aware Policy}},
  author        = {Myungkyu Koo and Daewon Choi and Taeyoung Kim and Kyungmin Lee and Changyeon Kim and Younggyo Seo and Jinwoo Shin},
  booktitle     = {The Fourteenth International Conference on Learning Representations},
  year          = {2026},
  url           = {https://openreview.net/forum?id=KcJ9U0x6kO},
}

@inproceedings{kim2026cosmospolicy,
  title         = {{Cosmos Policy: Fine-Tuning Video Models for Visuomotor Control and Planning}},
  author        = {Kim, Moo Jin and Gao, Yihuai and Lin, Tsung-Yi and Lin, Yen-Chen and Ge, Yunhao and Lam, Grace and Liang, Percy and Song, Shuran and Liu, Ming-Yu and Finn, Chelsea and others},
  booktitle     = {International Conference on Learning Representations},
  year          = {2026},
  volume        = {2026},
  pages         = {71531--71552},
  editor        = {C. Vondrick and B. Hariharan and C. Raffel and L. Pinto and D. Yang and A. Faust},
  url           = {https://proceedings.iclr.cc/paper_files/paper/2026/file/748becc400a57c0e31cfe6a2e7951467-Paper-Conference.pdf},
}

@inproceedings{yang2026cowvla,
  title         = {{Chain of World: World Model Thinking in Latent Motion}},
  author        = {Yang, Fuxiang and Di, Donglin and Tang, Lulu and Zhang, Xuancheng and Fan, Lei and Li, Hao and Chen, Wei and Su, Tonghua and Ma, Baorui},
  booktitle     = {Proceedings of the IEEE/CVF Conference on Computer Vision and Pattern Recognition (CVPR)},
  year          = {2026},
  pages         = {6675--6684},
  month         = {June},
}

@inproceedings{zhong2026acotvla,
  title         = {{ACoT-VLA: Action Chain-of-Thought for Vision-Language-Action Models}},
  author        = {Zhong, Linqing and Liu, Yi and Wei, Yifei and Xiong, Ziyu and Liu, Si and Ren, Guanghui},
  booktitle     = {Proceedings of the IEEE/CVF Conference on Computer Vision and Pattern Recognition (CVPR)},
  year          = {2026},
  pages         = {8152--8162},
  month         = {June},
}

@inproceedings{zhang2026atomicvla,
  title         = {{AtomicVLA: Unlocking the Potential of Atomic Skill Learning in Robots}},
  author        = {Zhang, Likui and Tang, Tao and Zhan, Zhihao and Chen, Xiuwei and Chen, Zisheng and Han, Jianhua and Zhu, Jiangtong and Xu, Pei and Xu, Hang and Wu, Hefeng and others},
  booktitle     = {Proceedings of the IEEE/CVF Conference on Computer Vision and Pattern Recognition (CVPR)},
  year          = {2026},
  pages         = {20743--20754},
  month         = {June},
}

@inproceedings{hu2026behaviorvla,
  title         = {{From Abstraction to Instantiation: Learning Behavioral Representation for Vision-Language-Action Model}},
  author        = {Hu, Bing and Li, Zaijing and Shao, Rui and Chen, Junda and Liu, April Hua and Zheng, Wei-Shi and Nie, Liqiang},
  booktitle     = {Proceedings of the 43rd International Conference on Machine Learning},
  year          = {2026},
  volume        = {306},
  pages         = {45272--45289},
  month         = {06--11 Jul},
  editor        = {Zhang, Tong and Dudik, Miroslav and Jaggi, Martin and Agarwal, Alekh and Li, Sharon and Schuurmans, Dale and Zhu, Jerry and Berkenkamp, Felix and Dong, Hanze and Bietti, Alberto},
  series        = {Proceedings of Machine Learning Research},
  publisher     = {PMLR},
  url           = {https://proceedings.mlr.press/v306/hu26ad.html},
}

@inproceedings{liu2026streampi,
  title         = {{StreamPI: Streaming Multimodal Temporal Modeling for Vision-Language-Action Models}},
  author        = {Liu, Zhe and Hou, Jinghua and Lu, Yuxiang and Yang, Zhenya and Fan, Xianzhe and Luo, Junwei and Li, Junyi and Han, Ruihua and Hou, Zhi and Zhao, Hengshuang},
  booktitle     = {Advances in Neural Information Processing Systems},
  year          = {2026},
}

@inproceedings{jang2026mgselect,
  title         = {{Verifier-free Test-Time Sampling for Vision-Language-Action Models}},
  author        = {Suhyeok Jang and Dongyoung Kim and Changyeon Kim and Youngsuk Kim and Jinwoo Shin},
  booktitle     = {The Fourteenth International Conference on Learning Representations},
  year          = {2026},
  url           = {https://openreview.net/forum?id=UD4Rw8MOEK},
}

@inproceedings{zhang2026vlm4vla,
  title         = {{{VLM}4{VLA}: Revisiting Vision-Language-Models in Vision-Language-Action Models}},
  author        = {Jianke Zhang and Xiaoyu Chen and Yanjiang Guo and Yucheng Hu and Jianyu Chen},
  booktitle     = {The Fourteenth International Conference on Learning Representations},
  year          = {2026},
  url           = {https://openreview.net/forum?id=tc2UsBeODW},
}

@inproceedings{jin2026structvla,
  title         = {{Beyond Dense Futures: World Models as Structured Planners for Robotic Manipulation}},
  author        = {Jin, Minghao and Liao, Mozheng and Han, Mingfei and Li, Zhihui and Chang, Xiaojun},
  booktitle     = {Computer Vision -- ECCV 2026},
  year          = {2026},
  pages         = {378--396},
  editor        = {Favaro, Paolo and Kukelova, Zuzana and Maki, Atsuto and Rohrbach, Anna and Schindler, Konrad and Tombari, Federico},
  publisher     = {Springer Nature Switzerland},
  address       = {Cham},
  isbn          = {978-3-032-37138-6},
}

@inproceedings{li2026spvla,
  title         = {{SP-VLA: A Joint Model Scheduling and Token Pruning Approach for VLA Model Acceleration}},
  author        = {Li, Ye and Meng, Yuan and Sun, Zewen and Ji, Kangye and Tang, Chen and Fan, Jiajun and Ma, Xinzhu and Xia, Shu-Tao and Wang, Zhi and Zhu, Wenwu},
  booktitle     = {International Conference on Learning Representations},
  year          = {2026},
  volume        = {2026},
  pages         = {38155--38184},
  editor        = {C. Vondrick and B. Hariharan and C. Raffel and L. Pinto and D. Yang and A. Faust},
  url           = {https://proceedings.iclr.cc/paper_files/paper/2026/file/4072543747a14bbed76284cf2c04b9e9-Paper-Conference.pdf},
}

@article{community2026starvla,
  title         = {{StarVLA: A Lego-like Codebase for Vision-Language-Action Model Developing}},
  author        = {{StarVLA Community}},
  journal       = {arXiv preprint arXiv:2604.05014},
  year          = {2026},
  eprint        = {2604.05014},
  archivePrefix = {arXiv},
  primaryClass  = {cs.RO},
  url           = {https://arxiv.org/abs/2604.05014},
}

@article{bai2025qwen25vl,
  title         = {{Qwen2.5-VL Technical Report}},
  author        = {Shuai Bai and Keqin Chen and Xuejing Liu and Jialin Wang and Wenbin Ge and Sibo Song and Kai Dang and Peng Wang and Shijie Wang and Jun Tang and others},
  journal       = {arXiv preprint arXiv:2502.13923},
  year          = {2025},
  eprint        = {2502.13923},
  archivePrefix = {arXiv},
  primaryClass  = {cs.CV},
  url           = {https://arxiv.org/abs/2502.13923},
}

@inproceedings{wang2025vggt,
  title         = {{VGGT: Visual Geometry Grounded Transformer}},
  author        = {Wang, Jianyuan and Chen, Minghao and Karaev, Nikita and Vedaldi, Andrea and Rupprecht, Christian and Novotny, David},
  booktitle     = {Proceedings of the IEEE/CVF Conference on Computer Vision and Pattern Recognition (CVPR)},
  year          = {2025},
  pages         = {5294--5306},
  month         = {June},
}

@inproceedings{ma2025gloverpp,
  title         = {{GLOVER++: Unleashing the Potential of Affordance Learning from Human Behaviors for Robotic Manipulation}},
  author        = {Ma, Teli and Zheng, Jia and Wang, Zifan and Gao, Ziyao and Zhou, Jiaming and Liang, Junwei},
  booktitle     = {Proceedings of The 9th Conference on Robot Learning},
  year          = {2025},
  volume        = {305},
  pages         = {3972--3994},
  month         = {27--30 Sep},
  editor        = {Lim, Joseph and Song, Shuran and Park, Hae-Won},
  series        = {Proceedings of Machine Learning Research},
  publisher     = {PMLR},
  url           = {https://proceedings.mlr.press/v305/ma25b.html},
}

@article{hu2024metric3dv2,
  title         = {{Metric3D v2: A Versatile Monocular Geometric Foundation Model for Zero-Shot Metric Depth and Surface Normal Estimation}},
  author        = {Hu, Mu and Yin, Wei and Zhang, Chi and Cai, Zhipeng and Long, Xiaoxiao and Chen, Hao and Wang, Kaixuan and Yu, Gang and Shen, Chunhua and Shen, Shaojie},
  journal       = {IEEE Transactions on Pattern Analysis and Machine Intelligence},
  year          = {2024},
  volume        = {46},
  number        = {12},
  pages         = {10579--10596},
  doi           = {10.1109/TPAMI.2024.3444912},
}

@article{oquab2024dinov2,
  title         = {{{DINO}v2: Learning Robust Visual Features without Supervision}},
  author        = {Maxime Oquab and Timoth{\'e}e Darcet and Th{\'e}o Moutakanni and Huy V. Vo and Marc Szafraniec and Vasil Khalidov and Pierre Fernandez and Daniel Haziza and Francisco Massa and Alaaeldin El-Nouby and others},
  journal       = {Transactions on Machine Learning Research},
  year          = {2024},
  issn          = {2835-8856},
  url           = {https://openreview.net/forum?id=a68SUt6zFt},
}

@inproceedings{liu2023libero,
  title         = {{LIBERO: Benchmarking Knowledge Transfer for Lifelong Robot Learning}},
  author        = {Liu, Bo and Zhu, Yifeng and Gao, Chongkai and Feng, Yihao and Liu, Qiang and Zhu, Yuke and Stone, Peter},
  booktitle     = {Advances in Neural Information Processing Systems},
  year          = {2023},
  volume        = {36},
  pages         = {44776--44791},
  editor        = {A. Oh and T. Naumann and A. Globerson and K. Saenko and M. Hardt and S. Levine},
  publisher     = {Curran Associates, Inc.},
  doi           = {10.52202/075280-1939},
  url           = {https://proceedings.neurips.cc/paper_files/paper/2023/file/8c3c666820ea055a77726d66fc7d447f-Paper-Datasets_and_Benchmarks.pdf},
}

@article{mees2022calvin,
  title         = {{CALVIN: A Benchmark for Language-Conditioned Policy Learning for Long-Horizon Robot Manipulation Tasks}},
  author        = {Mees, Oier and Hermann, Lukas and Rosete-Beas, Erick and Burgard, Wolfram},
  journal       = {IEEE Robotics and Automation Letters},
  year          = {2022},
  volume        = {7},
  number        = {3},
  pages         = {7327--7334},
  doi           = {10.1109/LRA.2022.3180108},
}

@inproceedings{li2024simpler,
  title         = {{Evaluating Real-World Robot Manipulation Policies in Simulation}},
  author        = {Li, Xuanlin and Hsu, Kyle and Gu, Jiayuan and Mees, Oier and Pertsch, Karl and Walke, Homer Rich and Fu, Chuyuan and Lunawat, Ishikaa and Sieh, Isabel and Kirmani, Sean and others},
  booktitle     = {Proceedings of The 8th Conference on Robot Learning},
  year          = {2025},
  volume        = {270},
  pages         = {3705--3728},
  month         = {06--09 Nov},
  editor        = {Agrawal, Pulkit and Kroemer, Oliver and Burgard, Wolfram},
  series        = {Proceedings of Machine Learning Research},
  publisher     = {PMLR},
  url           = {https://proceedings.mlr.press/v270/li25c.html},
}

@inproceedings{go2026flowvla,
  title         = {{FlowVLA}: Enabling World Dynamics Understanding through Pluggable {3D} Flow Learning},
  author        = {Go, Junhyeong and Jung, Mingyu and Im, Woobin and Ryu, Jongbin},
  booktitle     = {Conference on Robot Learning},
  year          = {2026},
  url           = {https://arxiv.org/abs/2610.16160},
}
}


\end{document}